\documentclass[english]{article}
\usepackage[T1]{fontenc}
\usepackage[margin=1.1in,letterpaper]{geometry}

\usepackage{lipsum}
\usepackage{amsfonts}
\usepackage{graphicx}
\usepackage{epstopdf}
\usepackage{algorithmic}

\usepackage{enumitem}
\setlist[enumerate]{leftmargin=.5in}
\setlist[itemize]{leftmargin=.5in}

\usepackage{amsthm}
\newtheoremstyle{remarkstyle} 
  {6pt}   
  {6pt}   
  {\normalfont} 
  {}      
  {\itshape} 
  {.}     
  { }     
  {}      
  
\theoremstyle{remarkstyle}
\newtheorem*{remark}{Remark}

\usepackage{mathtools} 
\usepackage{booktabs} 
\usepackage{tikz} 

\usepackage{microtype}
\usepackage{graphicx}
\usepackage{subfigure}
\usepackage{epstopdf}
\usepackage{hyperref}
\usepackage{amsmath}
\usepackage{amsfonts}
\usepackage[utf8]{inputenc}
\graphicspath{{figs}}
 
\usepackage{amsmath}
\usepackage{algorithm,algorithmic}
\usepackage{amssymb}
\usepackage{mathtools}
\usepackage{tikz}
\usetikzlibrary{bayesnet}
\usetikzlibrary{arrows}
\usetikzlibrary{positioning}
\usetikzlibrary{backgrounds}
\usepackage{xcolor,soul}
\usepackage{bm}
\usepackage{bbm}
\usepackage{wasysym}

\def\bydef{\triangleq}

\title{PCE-Net: High Dimensional Surrogate Modeling for Learning Uncertainty}

\author{Paz Fink Shustin\thanks{University of Oxford, UK}\and 
Shashanka Ubaru\thanks{IBM Research, USA}\and 
Ma{\l}gorzata J. Zimo\'{n}\thanks{IBM Research Europe, Daresbury, UK and Department of Mathematics, University of Manchester, UK}\and 
Songtao Lu\footnotemark[2]\and 
Vasileios Kalantzis\footnotemark[2]\and 
Lior Horesh\footnotemark[2]\and 
Haim Avron\thanks{Tel-Aviv University, Israel}}

\usepackage{amsopn}

\date{}

\begin{document}
\newcommand{\haim}[1]{\textcolor{red}{[Haim: #1]}}

\newcommand{\paz}[1]{\textcolor{magenta}{[Paz: #1]}}

\global\long\def\conj{*}%

\global\long\def\Z{\mathbb{Z}}%

\global\long\def\R{\mathbb{R}}%

\global\long\def\C{\mathbb{C}}%

\global\long\def\H{{\cal H}}%

\global\long\def\X{{\cal X}}%

\global\long\def\Q{{\cal Q}}%

\global\long\def\Y{{\cal Y}}%

\global\long\def\e{{\mathbf{e}}}%

\global\long\def\et#1{{\e(#1)}}%

\global\long\def\ef{{\mathbf{\et{\cdot}}}}%

\global\long\def\x{{\mathbf{x}}}%

\global\long\def\w{{\mathbf{w}}}%

\global\long\def\xt#1{{\x(#1)}}%

\global\long\def\xf{{\mathbf{\xt{\cdot}}}}%

\global\long\def\d{{\mathbf{d}}}%

\global\long\def\b{{\mathbf{b}}}%

\global\long\def\u{{\mathbf{u}}}%

\global\long\def\y{{\mathbf{y}}}%

\global\long\def\yt#1{{\y(#1)}}%

\global\long\def\yf{{\mathbf{\yt{\cdot}}}}%

\global\long\def\z{{\mathbf{z}}}%

\global\long\def\v{{\mathbf{v}}}%

\global\long\def\h{{\mathbf{h}}}%

\global\long\def\s{{\mathbf{s}}}%

\global\long\def\c{{\mathbf{c}}}%

\global\long\def\p{{\mathbf{p}}}%

\global\long\def\f{{\mathbf{f}}}%

\global\long\def\g{{\mathbf{g}}}%

\global\long\def\a{{\mathbf{a}}}%

\global\long\def\rb{{\mathbf{r}}}%

\global\long\def\rt#1{{\rb(#1)}}%

\global\long\def\rf{{\mathbf{\rt{\cdot}}}}%

\global\long\def\mat#1{{\ensuremath{\bm{\mathrm{#1}}}}}%

\global\long\def\valpha{\mat{\alpha}}%

\global\long\def\vbeta{\mat{\beta}}%

\global\long\def\vtheta{\mat{\theta}}%

\global\long\def\veta{\mat{\eta}}%

\global\long\def\vnu{\mat{\nu}}%

\global\long\def\vmu{\mat{\mu}}%

\global\long\def\vsigma{\mat{\sigma}}%

\global\long\def\vrho{\mat{\rho}}%

\global\long\def\vphi{\mat{\phi}}%

\global\long\def\matN{\ensuremath{{\bm{\mathrm{N}}}}}%

\global\long\def\matA{\ensuremath{{\bm{\mathrm{A}}}}}%

\global\long\def\matB{\ensuremath{{\bm{\mathrm{B}}}}}%

\global\long\def\matC{\ensuremath{{\bm{\mathrm{C}}}}}%

\global\long\def\matD{\ensuremath{{\bm{\mathrm{D}}}}}%

\global\long\def\matP{\ensuremath{{\bm{\mathrm{P}}}}}%

\global\long\def\matU{\ensuremath{{\bm{\mathrm{U}}}}}%

\global\long\def\matV{\ensuremath{{\bm{\mathrm{V}}}}}%

\global\long\def\matM{\ensuremath{{\bm{\mathrm{M}}}}}%

\global\long\def\matR{\mat R}%

\global\long\def\matW{\mat W}%

\global\long\def\matK{\mat K}%

\global\long\def\matQ{\mat Q}%

\global\long\def\matS{\mat S}%

\global\long\def\matY{\mat Y}%

\global\long\def\matX{\mat X}%

\global\long\def\matI{\mat I}%

\global\long\def\matJ{\mat J}%

\global\long\def\matZ{\mat Z}%

\global\long\def\matL{\mat L}%

\global\long\def\S#1{{\mathbb{S}_{N}[#1]}}%

\global\long\def\IS#1{{\mathbb{S}_{N}^{-1}[#1]}}%

\global\long\def\PN{\mathbb{P}_{N}}%

\global\long\def\TNormS#1{\|#1\|_{2}^{2}}%

\global\long\def\TNorm#1{\|#1\|_{2}}%

\global\long\def\InfNorm#1{\|#1\|_{\infty}}%

\global\long\def\FNorm#1{\|#1\|_{F}}%

\global\long\def\UNorm#1{\|#1\|_{\matU}}%

\global\long\def\UNormS#1{\|#1\|_{\matU}^{2}}%

\global\long\def\UINormS#1{\|#1\|_{\matU^{-1}}^{2}}%

\global\long\def\ANorm#1{\|#1\|_{\matA}}%

\global\long\def\BNorm#1{\|#1\|_{\mat B}}%

\global\long\def\HNormS#1{\|#1\|_{\H}^{2}}%

\global\long\def\HNorm#1{\|#1\|_{\H}}%

\global\long\def\XNormS#1#2{\|#1\|_{#2}^{2}}%

\global\long\def\AINormS#1{\|#1\|_{\matA^{-1}}^{2}}%

\global\long\def\BINormS#1{\|#1\|_{\matB^{-1}}^{2}}%

\global\long\def\BINorm#1{\|#1\|_{\matB^{-1}}}%

\global\long\def\ONorm#1#2{\|#1\|_{#2}}%

\global\long\def\T{\textsc{T}}%

\global\long\def\pinv{\textsc{+}}%

\global\long\def\Expect#1{{\mathbb{E}}\left[#1\right]}%

\global\long\def\ExpectC#1#2{{\mathbb{E}}_{#1}\left[#2\right]}%

\global\long\def\dotprod#1#2#3{(#1,#2)_{#3}}%

\global\long\def\dotprodsqr#1#2{(#1,#2)^{2}}%

\global\long\def\Trace#1{{\bf Tr}\left(#1\right)}%

\global\long\def\realpart#1{{\bf Re}\left(#1\right)}%

\global\long\def\nnz#1{{\bf nnz}\left(#1\right)}%

\global\long\def\range#1{{\bf range}\left(#1\right)}%

\global\long\def\nully#1{{\bf null}\left(#1\right)}%

\global\long\def\vecmat#1{{\bf vec}\left(#1\right)}%

\global\long\def\vol#1{{\bf vol}\left(#1\right)}%

\global\long\def\rank#1{{\bf rank}\left(#1\right)}%

\global\long\def\diag#1{{\bf diag}\left(#1\right)}%

\global\long\def\erfc#1{{\bf erfc}\left(#1\right)}%

\global\long\def\grad#1{{\bf grad}#1}%

\global\long\def\st{\,\,\,\text{s.t.}\,\,\,}%

\global\long\def\KL#1#2{D_{{\bf KL}}\left(#1,#2\right)}%

\maketitle

\begin{abstract}
  Learning data representations under uncertainty is an important task that emerges in numerous scientific computing and data analysis applications. However, uncertainty quantification techniques are computationally intensive and become prohibitively expensive for high-dimensional data.
In this study, we introduce a dimensionality reduction surrogate modeling (DRSM) approach for representation learning and uncertainty quantification that aims to deal with data of moderate to high dimensions. 
The approach involves a two-stage learning process: 1) employing a {\em variational autoencoder} to learn a low-dimensional representation of the input data distribution; and 2) harnessing {\em polynomial chaos expansion (PCE)} formulation to map the low dimensional distribution to the output target. 
The model enables us to (a) capture the system dynamics efficiently in the low-dimensional latent space, (b) learn under uncertainty, a representation of the data and a mapping between input and output distributions, (c) estimate this uncertainty in the high-dimensional data system, and (d) match high-order moments of the output distribution; without any prior statistical assumptions on the data. Numerical results are presented to illustrate the performance of the proposed method.
\end{abstract}

\newcommand{\keywords}[1]{\par\noindent\textbf{Keywords:} #1}
\newcommand{\msc}[1]{\par\noindent\textbf{MSC codes:} #1}
\msc{68T01, 68T05, 37M99, 65C99, 65P20 }

\section{Introduction}\label{sec:intro}
Learning the input-output (I/O) relations of a given data system is a fundamental problem that occurs in several applications including supervised learning, solving and learning partial differential equations (PDEs), control systems, signal processing, computer vision, natural language processing, and many more~\cite{jordan2015machine}. In recent times, neural-networks (NNs) have been popularly employed for this purpose, and have been shown to be comprehensive and highly effective in these applications~\cite{goodfellow2016deep,han2018solving,deng2018deep,fadlullah2017state,voulodimos2018deep}. In many situations, the tasks of learning data representation and I/O relationship also needs to account for the uncertainty in the data. 
Thus, the learning model should also offer means to perform uncertainty quantification (UQ)~\cite{smith2013uncertainty}.
For example, uncertainty in the data (also known as aleatoric uncertainty) arises due to reasons such as noise, training, and testing data mismatch, incomplete data, class overlap,  multi-modal data, and others~\cite{malinin2019uncertainty,abdar2021review, hullermeier2021aleatoric}. Complementarily, uncertainty in the model/system (i.e., epistemic uncertainty) occurs due to inadequate knowledge, incorrect assumptions upon data distributions and/or model functions, natural variability in system parameters, faulty sub-systems, and more~\cite{abdar2021review,schobi2015polynomial, hullermeier2021aleatoric}.

Given the importance of the problem, numerous methods for uncertainty modeling and quantification have been proposed in different engineering fields~\cite{smith2013uncertainty,sullivan2015introduction} and in the artificial intelligence literature~\cite{abdar2021review}. Traditional UQ techniques are typically stochastic sampling-based simulation methods~\cite{mohamed2010comparison}. However, these methods are computationally expensive, making them inapplicable to modern large and complex data models. 
Alternatively, surrogate modeling (also known as response surface or meta-model) techniques such as Polynomial Chaos Expansions (PCEs) \cite{ ghanem1990polynomial, xiu2002wiener, soize2004physical}, Gaussian Process (GP) modeling and regression~\cite{rasmussen2003gaussian,chen2015uncertainty}, and support vector machines~\cite{li2006support} have received much attention due to their low computational cost. Recently, deep learning methods~\cite{tripathy2018deep,zhang2019quantifying,zheng2021mini}, including Bayesian neural networks~\cite{wang2016towards,abdar2021review} have been used as  surrogate models for UQ. However, parameterizing and training most of these surrogate models will be intractable when the number of input parameters becomes large (known as the "curse of dimensionality"~\cite{verleysen2005curse}), i.e., for high-dimensional data systems.

 \begin{figure}[t!]
\centering{}
 \includegraphics[clip,scale=0.8]{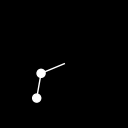} \includegraphics[clip,scale=0.8]{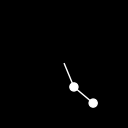}
  \includegraphics[clip,scale=0.8]{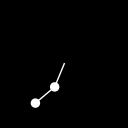} \includegraphics[clip,scale=0.8]{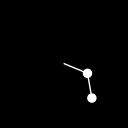}
   \includegraphics[clip,scale=0.8]{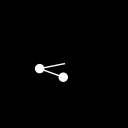}
\caption{Example images of a double pendulum at five different random states.}\label{fig:double_pendulum}
\end{figure}

In numerous situations, even though the data observations (descriptive features) of the given system are high-dimensional, the intrinsic dimensionality of the system and the corresponding number of hidden state variables can be quite low. As a motivating example, we consider the double pendulum problem studied in~\cite{chen2022automated}; see Figure~\ref{fig:double_pendulum}. Here, a swinging double pendulum is observed as a set of $128\times 128$ image frames (16,384-dimensional inputs).  However, we know that the state and dynamics of the pendulum can be fully
described using just four variables, namely the two angles and two angular velocities (additional details and results are presented in Section~\ref{uq_in_experiments}). In such situations, dimensionality reduction approaches~\cite{ghanem2003stochastic,ma2011kernel,mehrasa2019variational,chen2022automated, kim2024dimensionality} and dimensionality reduction
surrogate modeling (DRSM) methods~\cite{tripathy2016gaussian,lataniotis2020extending} can be employed to model the system dynamics and predictions, learning representations and UQ. However, for efficient learning and UQ, it is imperative that the dimensionality reduction approach used can effectively capture the system dynamics in low-dimensional space. 

 In this paper, we study a surrogate modeling approach for dimensionality reduction (which we call  \texttt{PCE-Net}) to learn representations of high-dimensional data systems under uncertainty. The approach learns the functional mapping between the input and output data distributions, where both distributions could be unknown a priori. 
This assists in data uncertainty modeling and propagation, as well as in promoting generalizability~\cite{wilson2020bayesian}. The proposed approach comprises of two stages. In the first stage, we map the (possibly high-dimensional) input data distribution to a low-dimensional latent distribution via NNs. In the second stage,  a surrogate model is trained to learn a mapping between the latent and output distributions.
For the dimensionality reduction stage, we employ variational autoencoders (VAE)~\cite{kingma2013auto}, a NN-based Bayesian unsupervised learning approach. VAE embeds the (possibly unknown) input data distribution to a normal distribution in a lower-dimensional latent space, enabling us to use a suitable surrogate model to map the latent space to the output. As a non-linear approach, VAE enables us to capture the system/data dynamics efficiently in the latent space. Moreover, VAE is a  Bayesian approach, and it has recently been used for input data uncertainty quantification~\cite{bohm2019uncertainty,mehrasa2019variational,guo2020deep}. 

In the surrogate modeling stage, we consider PCEs \cite{xiu2002wiener, ghanem1990polynomial, oladyshkin2012data} to learn the mapping from the latent distribution to the output space. PCEs are highly efficient uncertainty modeling techniques and have many appealing properties, including: (a) they are inexpensive to compute when the input dimension is small; (b) they can match higher-order moments, making them suitable for arbitrary distributions with arbitrary probability measures~\cite{oladyshkin2012data}; and (c) they capture the global characteristic of the function~\cite{schobi2015polynomial}. In our study, we also explore a loss function with a {\em maximum mean discrepancy} (MMD) based regularization for computing the PCE coefficients, and a bilevel alternating minimization (BAM) procedure~\cite{xiao2023alternating,xiao2023generalized} to {\em jointly} solve for the PCE coefficients and the hyperparameters of the MMD regularization. 
The approach only requires sampling the latent distribution and does not require any prior statistical assumptions on the data.

In the learning set-up, we are given a dataset $(\x_{1},y_{1}),\ldots,(\x_{n},y_{n})$ where  $\{\x_i\}_{i=1}^n \in \mathcal{X}\subseteq \mathbb{R}^{d}$ and $\{y_i\}_{i=1}^n \in \mathcal{Y}\subseteq \mathbb{R}$, and a function $F:\mathcal{X}\longrightarrow \mathcal{Y}$,
such that the observations satisfy $y_i = F(\x_i)$.  We assume that $F$ is unknown and expensive to evaluate. Hence, our goal is to build a cheap-to-compute function $\tilde{F}_{\vrho}:\mathbb{R}^{d}\to\mathbb{R}$,  parameterized  by  $\vrho$ (that may depend on the priors of $\{\mathcal{X},\mathcal{Y\}}$), that approximates $F$ well with respect to certain metric, i.e., $F(\x_i) = \tilde{F}_{\vrho}(\x_i)+\varepsilon$, where $\varepsilon$ refers to an error term.
First, we use a dimensionality
reduction function $E:\mathbb{R}^{d}\to\mathbb{R}^{m}$ that maps
$\{\x_{1},\dots,\x_{n}\}$ to $\{\z_{1},\dots,\z_{n}\}$, where each $\z_i$'s have separate normal distributions $\mathcal{N}(\vmu_i,\mat{\Sigma}_i)$ .
Here, we set the latent dimension (a hyperparameter) $m$ to be much smaller than the input dimension $d$. The function
$E$ will hence be the encoder part of VAE (see Section~\ref{ssec:VAE} for a review). Second, in order to map $\{\z_{1},\dots,\z_{n}\}$ to $\{y_1,\dots,y_n\}$, we use a PCE function $P:\R^{m}\to\R$ with  expansion form:
\[
P(\mat{\z})=\sum_{j=1}^{\ell}c_{j}\varphi_{j}(\mat{\z}) \, , 
\]
where we construct a family
of multivariate polynomials $\left\{ \varphi_{j}\right\}_{j=1}^{\ell}$ that are orthogonal w.r.t. the prior distribution of $\z$, and the coefficients $\left\{ c_{j}\right\}_{j=1}^{\ell}$ are learned using 
an $L_2$ loss function.  We also explore the benefits of using an MMD-based regularization for learning these PCE coefficients.
Therefore, the proposed surrogate model can simply be written as $\tilde{F}(\x)=P\circ E(\x)=P(\mat{\z})$.


 \paragraph{Outline} In Section~\ref{sec:prelims}, we provide details on the background information required for our study. We also discuss some of the prior works that are closely related to the approach. In Section~\ref{sec:pcenet}, we present the \texttt{PCE-Net}  method for learning data representation under uncertainty and discuss different aspects and characteristics of the method. Numerical results on multiple datasets from different applications, including the double pendulum problem, illustrating the performance of \texttt{PCE-Net} are presented in Section~\ref{sec:expts}.



\section{Preliminaries}\label{sec:prelims}

In this section, we first present the notation details, and then discuss the two main ingredients of \texttt{PCE-Net}, namely,  Variational Autoencoder (VAE) and Polynomial Chaos Expansion (PCE). We also briefly discuss the Maximum Mean Discrepancy (MMD) loss function. We end the section with a discussion on some of  the related prior works.

\paragraph{ Notation}
We will follow the standard notation of lowercase, bold lowercase, and bold uppercase letters for scalars $x$, vectors $\x$, and matrices $\matX$, respectively. Probability vector spaces are denoted using calligraphic letters $\mathcal{X}$ and functions using uppercase letters $F(\cdot)$. 
$D_{\mathrm{KL}}$ denotes the Kullback-Leibler divergence (KL-divergence), and $f_\mat{X}(\cdot)$ denotes the joint probability density function with $\matX$. Finally, $\mathcal{N} \left( \vmu , \mat{\Sigma} \right)$ denotes the multivariate Gaussian distribution with mean vector $\vmu$ and covariance matrix~$\mat{\Sigma}$.

\subsection{Variational Autoencoder}\label{ssec:VAE}
VAE was first introduced in \cite{kingma2013auto}, and is a Bayesian unsupervised learning technique based on dimensionality reduction and variational inference (VI)~\cite{hinton1993keeping,waterhouse1996bayesian,jordan1998introduction}. VAE comprises two parts. The first part is the encoder parameterized by $\vphi$ which takes input $\x\in \mathbb{R}^d$ and returns a distribution on the latent variable $\z\in \mathbb{R}^m$, where $m<d$. The second part is the decoder parametrized by $\vtheta$ which tries to reconstruct $\x$ from the samples of the latent distribution. Together with VI, the encoder is the inference model and the decoder is the generative model. These two parts (NNs) are jointly optimized in order to maximize  the evidence lower bound (ELBO):
\begin{align} \label{ELBO}
\begin{split}
    \cal{L} \left( \vtheta, \vphi; \x \right) & = \mathbb{E}_q \left[  \ln p_{\vtheta}(\x,\z) - \ln q_{\vphi}(\z\mid\x) \right] \\
    & = \ln p_{\vtheta}(\x) - D_{\mathrm{KL}} \left[ q_{\vphi}(\z\mid\x) || p(\z\mid\x) \right].
\end{split}
\end{align}
Here, an intractable posterior distribution $p_{\vtheta}(\z\mid\x)$ of a latent variable model $p_{\vtheta}(\x,\z)=p_{\vtheta}(\x\mid\z)p_{\vtheta}(\z)$ is approximated by a guide $q_{\vphi}(\z\mid\x)$. The approximation, $q_{\vphi}(\z\mid\x)$, is performed by taking $q_{\vphi}(\z\mid\x)$ to be a simple distribution, e.g., Gaussian with a diagonal covariance $\cal{N} \left( \vmu(\x), \mat{\Sigma}(\x) \right)$. The parameters of $q_{\vphi}(\z\mid\x)$ are estimated by maximizing Eq.~\eqref{ELBO}.
In VAE, the encoder outputs $q_{\vphi}(\z\mid\x)$ by returning $\vmu(\x)$ and (the diagonal elements of) $\mat{\Sigma}(\x)$, and then $\z$ is sampled and passed through the decoder which allows the optimization of the ELBO. Importantly, ELBO can also be written in the form: 
\begin{equation*} \label{ELBO_2}
    \mathcal{L} \left( \vtheta,\vphi ; \x \right) = \mathbb{E}_q \left[  \ln p_{\vtheta}(\x\mid\z) \right] - D_{\mathrm{KL}} \left[ q_{\vphi}(\z\mid\x) || p_{\vtheta}(\z) \right] \, ,
\end{equation*}
where $\ln p_{\vtheta}(\x\mid\z)$ is the marginal log-likelihood. The term $D_{\mathrm{KL}} \left[ q_{\vphi}(\z\mid\x) || p_{\vtheta}(\z) \right]$ can be viewed as a regularization term which forces $q_{\vphi}(\z\mid\x)$ to be approximately distributed as the prior $p_{\vtheta}(\z)$, which is independent of $\x$. Thus, $q_{\vphi}(\z)$ is approximately distributed as the prior. Typically, the distributions $q_{\vphi}(\z\mid\x)$ and $p_{\vtheta}(\z)$ are chosen to be Gaussians.

\paragraph{Learning disentanglement}\label{par:disentanglement} $\beta$-VAE, originally proposed in~\cite{higgins2017beta}, is designed to learn independent generative factors of a dataset in an unsupervised manner. In $\beta$-VAE,  the KL-divergence term in the loss function is scaled to increase its influence. During the training of $\beta$-VAE, the following loss is used:
\begin{equation}
      \mathbb{E}_{q_{\phi}(\z\mid\x)} \left[\ln p_{\theta}(\x\mid\z) \right] - \beta D_{KL}\left(q_{\phi} (\z\mid\x) \big\| p_{\vtheta}(\z) \right).
\end{equation}
The first term corresponds to the reconstruction loss (data likelihood loss). The second term is the KL-divergence between the $q_{\phi} (\z\mid\x)$ values the network encodes for each $\x$, and a prior distribution $p_{\vtheta}(\z)$. Since KL-divergence is lowest when the two distributions are equivalent, this term pushes the latent $\z$ values to be more concentrated in the space of the prior multivariate Gaussian. This term is typically referred to as the regularization term and has been shown to learn latent embeddings that are disentangled~\cite{burgess2018understanding}. Further refinement of the work was suggested in~\cite{chen2018isolating}. The authors decomposed ELBO to show the existence of a term measuring the total correlation between latent variables. They scaled this component with $\beta$, introducing the $\beta$-TCVAE
(Total Correlation Variational Autoencoder) algorithm, providing a better trade-off between density estimation and disentanglement.

\subsection{Polynomial Chaos Expansion}\label{app:PCE}
PCE is an inexpensive surrogate modeling approach that aims to map uncertainty from an input space $\mathcal{Z} \subseteq \mathbb{R}^m$ to an output space $\mathcal{Y} \subseteq \mathbb{R}$. The uncertainty is expressed through a probabilistic framework using random vectors, i.e., $\mat{Y}=P(\mat{Z})$ where $\mat{Z}\in \cal{Z}$ with a given joint probability density function (PDF) $f_\mat{Z}(\cdot)$ and $P(\cdot)$ is a sum of polynomials that are typically orthogonal w.r.t. the measure $f_\mat{Z}$~\cite{ghanem1990polynomial, xiu2002wiener, soize2004physical}. In contrast to other probabilistic methods such as Gaussian Processes, PCEs approximate the global behavior of the model using a set of orthogonal polynomials. It is also assumed that $\mat{Y}$ has a finite variance $\mathbb{E}[\mat{Y}^2]<\infty$, and that each component of $\mat{Z}$ has finite moments of any order.

Thus, the space of square-integrable functions w.r.t. the weighted function $f_\mat{Z}(\cdot)$ can be represented by an orthonormal basis of polynomials $\{ \varphi_{\mat{i}} (\cdot) \}_{\mat{i}\in \mathbb{N}^d}$:
\begin{equation*} \label{PCE_ortho}
    \int_{\cal{Z}} \varphi_{\mat{i}} (\z) \varphi_{\mat{j}} (\z) f_\mat{Z}(\z) d\z = \delta_{\mat{i j}} \, .
\end{equation*}
Therefore, $\mat{Y}$ can be represented as
\begin{equation} \label{PCE_rep_inf}
    \mat{Y} = P(\mat{Z}) = \sum_{\mat{j}\in \mathbb{N}^d} c_{\mat{j}} \varphi_{\mat{j}} (\mat{Z}) \, .
\end{equation}
The coefficients $c_{\mat{j}}$ in \eqref{PCE_rep_inf} can be computed using a data driven regression approach~\cite{le2010spectral,schobi2015polynomial, torre2019data, lataniotis2020extending}. Given a dataset of observations $(\z_{1},y_{1}),\dots,(\z_{n},y_{n})\in \mathcal{Z}\times\mathcal{Y}$, the coefficients can be found by regression fitting, such as by minimizing the square loss function:
$
    \sum_{i=1}^{n} \left( y_i - P(\z_i) \right)^2.
$
Crucially, the orthonormality of the basis implies that the squared sum of the PCE coefficients typically displays rapid decay, which in turn reduces the number of coefficients actually required and avoids overfitting. 

Furthermore, in the case where the components of $\mat{Z}$ are independent and identically distributed, the polynomials in \eqref{PCE_rep_inf} are composed of univariate polynomials 
by a tensor product:
\begin{equation} \label{PCE_univ}
    \varphi_{\mat{j}} (\mat{Z}) = \prod_{k=1}^{d} \varphi_{j_k}^{(k)} (Z_k) \, ,
\end{equation}
where $\varphi_{j_k}^{(k)}$ is the $j_k$ polynomial in the $k$th dimension. Since using the series \eqref{PCE_rep_inf} is not practical,  PCEs are used as surrogate models which replace the true model in practice. This is done by truncating the series such that $|\mat{j}| = \sum_{k=1}^{d} j_k \leq \ell$:
\begin{equation*} \label{PCE_rep}
    P_{\ell}(\mat{Z}) = \sum_{j=1}^{\ell_p} c_{j} \varphi_{j} (\mat{Z}) \, ,
\end{equation*}
where  $\ell_p = \frac{(\ell+d)!}{\ell!d!}$.
For a large dimension $d$, the process becomes prohibitively expensive. 

\subsection{Maximum Mean Discrepancy}\label{MMD}
 In~\cite{gretton2012kernel},  a metric was presented for measuring distances between distributions in terms of mean embedding, which they termed maximum mean discrepancy (MMD).
Let $\cal{H}$ be a reproducing kernel Hilbert space (RKHS) over the domain $\cal{X}$ and $k: \cal{X}\times\cal{X} \to \mathbb{R}$ be the associated kernel. Denote $\vmu_{\veta} = \mathbb{E}_{x\sim \veta} [k(\x,\cdot)]$ as the kernel mean of a given probability measure $\veta$ over $\cal{X}$. Then, for two probability measures $\veta$ and $\vnu$ over $\cal{X}$, with mean embeddings $\vmu_{\veta}$ and $\vmu_{\vnu}$ respectively, the MMD is: 
$
\text{MMD} (\veta,\vnu)  = \HNormS{\vmu_{\veta} - \vmu_{\vnu}} \,
$
and can also be expressed as
\begin{align*}\text{MMD} (\veta,\vnu)    = \mathbb{E}_{\veta}  \left[ k(\x,\x') \right] - 2\mathbb{E}_{\veta,\vnu}  \left[ k(\x,\y) \right]   + \mathbb{E}_{\vnu} \left[ k(\y,\y') \right] \, ,
\end{align*}
where $\x,\x' \sim \veta$ and $\y,\y' \sim \vnu$. It can be seen that MMD is zero only if the two distributions are equal.

Given two sets of samples $\matX=\{\x_i\}_{i=1}^{n_1}$ and $\matY=\{\y_i \}_{i=1}^{n_2}$, one may ask whether their distributions $\veta$ and $\vnu$ are the same. For that purpose, an empirical estimate of MMD can be obtained by:
\begin{align}\label{emp_MMD}
{\cal L}_{{\text{MMD}}^2} & = \frac{1}{n_1^2} \sum_{i,j=1}^{{n_1}} k(\x_i,\x_j) - \frac{2}{{n_1}{n_2}} \sum_{i=1}^{n_1} \sum_{j=1}^{n_2} k(\x_i,\y_j)  + \frac{1}{n_2^2} \sum_{i,j=1}^{n_2} k(\y_i,\y_j) \, .
\end{align}

Note that, as a consequence, the resulting kernel mean may incorporate high-order moments of $\veta$. For example, when the kernel $k(\cdot,\cdot)$ is linear, $\mathbb{E}_{\veta}[k] = \vmu_{\veta}$, the mean of $\veta$. Choosing a Gaussian kernel allows us to capture all high-order moments, and MMD acts as a moment-matching approach;  see~\cite{li2015generative, kiasari2017generative} for details.


\subsection{Related Prior Work}
We discuss some of the prior work in the literature that are closely related to  \texttt{PCE-Net}. PCEs as surrogate models have been popularly used for data-driven uncertainty quantification and sensitivity analysis in numerous applications~\cite{xiu2002wiener,crestaux2009polynomial,najm2009uncertainty,sepahvand2010uncertainty, duong2016uncertainty, torre2019data,ubaru2021dynamic}. Arbitrary PCE~\cite{oladyshkin2012data} has been proposed to handle data with arbitrary and unknown distributions, and sparse PCE~\cite{blatman2011adaptive} was proposed for reducing the computational cost of PCEs. In \cite{schobi2015polynomial},  a method named PC-Krigging is proposed that combines PCEs with Gaussian Processes for improved global-local representation of the given data system. However, as previously mentioned, such surrogate modeling approaches are not applicable to high-dimensional data systems.

The idea of using dimensionality reduction (DR) methods for uncertainty quantification of high-dimensional data systems has been considered in the UQ literature~\cite{ghanem2003stochastic}, and DR methods such as principal component
analysis (PCA), kernel PCA~\cite{ma2011kernel}, active subspace methods~\cite{constantine2015exploiting} and autoencoders~\cite{mehrasa2019variational} have been used, also see~\cite{kim2024dimensionality}. Tripathy et al.~\cite{tripathy2016gaussian} proposed an active subspace approach (DR method) that is combined with a Gaussian Process (SM method) for high-dimensional uncertainty propagation. 
Later, in \cite{lataniotis2020extending}  a general framework for dimensionality reduction surrogate modelling (DRSM) approach for UQ is presented, and  various combinations of DR (PCA and kernel PCA) and SM (PCE and Gaussian processes) methods for UQ are studied. Their approach is to approximate the input data using a kernel density estimation, and then use a relative generalization error for learning the SM. Deep neural network based surrogate models have been proposed by~\cite{tripathy2018deep,zheng2021mini} for high-dimensional uncertainty quantification. Recently, a survey of methods for high-dimensional uncertainty quantification was presented in~\cite{kontolati2022survey}. However, the approach we study here differs from these methods in multiple aspects, namely: (a) the approach is to learn the mapping between input and output distributions, rather than point-wise fitting of the given training input and output data (hence, better generalization); (b) the distributions of the data in the latent space is directly learned and  kernel density estimation is not used (resulting in significant computational cost gain); (c) a Bayesian NN approach is used for DR, which capture system dynamics efficiently and helps in uncertainty quantification, and (d)  a moment matching approach is used to learn the coefficients of the SM (achieving improved output distribution matching).

\section{PCE-Net}\label{sec:pcenet}

In this section, we present the  \texttt{PCE-Net} method for high-dimensional uncertainty quantification. The method follows the DRSM approach and (a) uses VAE for learning a distribution of the input data in a low-dimensional latent space, and (b) then considers a PCE surrogate model to map the latent space to the output. 

We begin by training a VAE (or a $\beta$-VAE in cases where independent/disentangled latent space variables are required) on the given input data $\{\x_1,\dots,\x_n\}$. The learned parameters $\vphi$ of the encoder allow each of the data points $\x_i, \, i=1,\dots,n$ to be mapped to a distribution $\mathcal{N} \left( \vmu_i , \mat{\Sigma}_i \right)$, from which the corresponding $\z_i^{(j)} \in \mathbb{R}^m, \, j=1,\dots,n_s$ can be sampled for each $\x_i$'s (in the experiments, we consider $n_s\sim100$ samples). 
We denote by $E_{\vphi}(\cdot)$ the encoder and the sampling operations which map each data point $\x_i$ to  $\{\z_i^{(j)}\}_{j=1}^{n_s} \in \mathbb{R}^{m}$. The new set of data points $(\z_i^{(j)},y_i) \equiv \mathcal{D}_{\mathrm{tr}}$ is used in PCE learning, by considering the expectation of the response over the samples for each data point $i$. 
Therefore, the  PCE response for $\mathcal{N} \left( \vmu_i , \mat{\Sigma}_i \right)$ using the samples $\z_i^{(j)}$ is:
\begin{equation} \label{PCE_responses}
\tilde{y}_i = \frac{1}{n_s} \sum_{j=1}^{n_s} P_{\ell} (\z_i^{(j)}) = \frac{1}{n_s} \sum_{j=1}^{n_s} \sum_{k=1}^{\ell_p} c_{k} \varphi_{k} (\z_i^{(j)}) \, .
\end{equation}
We use  VAE prior joint distribution (univariate priors $\mathcal{N} \left( 0, 1 \right)$, for each of $\z_i^{(j)}$ i.i.d components) as the weight function for the PCE's Hermite polynomial basis, and thus we can utilize the tensorized form for PCE as given in Eq.~\eqref{PCE_univ}. The PCE coefficients  $c_k$'s can be computed by minimizing a least-squares  ($L_2$) loss function, as considered in our UQ study in Section~\ref{uq_in_experiments}.

\begin{figure}[tb!] \centering 
 \tikz{
 \node[obs] (y) {$y$};%
 \node[latent,below=of y, yshift=0.2cm] (zm) {$\hat{\z}$};
 \node[latent,left=of zm, xshift=0.2cm] (xt) {$\tilde{\x}$};
 \node[latent,left=of y, xshift=0.2cm] (z) {$\z$};
 \node[latent,left=of z, xshift=0.2cm, fill=gray!25] (x) {$\x$};
 \node[latent, above=of x, yshift=-0.2cm] (p) {$\vphi$};
 \node[latent, left=of z, xshift=-1.1cm, yshift=-0.5cm] (t) {$\vtheta$};
 \node[latent, below=of xt, yshift=0.2cm] (tt) {$\vphi$};
 \plate [inner sep=.25cm,yshift=.2cm] {plate1} {(x)(y)(z) (zm) (xt)} {$N$}

 \edge[dashed] {x} {z}
 \edge[dashed] {z} {xt}
 \edge[dashed] {xt} {zm}
 \edge {zm} {y} 
 \edge {z} {y}
 \edge {p} {z} 
 \edge {t} {xt} 
  \edge {tt} {zm} }
  \vskip -0.1in
 \caption{\texttt{PCE-Net} model. Latent variables $\z$ are used for $L_2$, and latent variables $\hat{\z}$ are used for MMD. } \label{graph_model_PCENet}
 \vskip-0.1in
\end{figure}

\paragraph{MMD Regularization} In addition to the least-squares approach, we also explore an alternate loss function to learn the PCE coefficients, where the square loss function is combined with the square root of the MMD loss function as a regularization term. This approach can be interpreted as aligning the distribution of the model responses with the distribution of empirical data, to capture the global structure of data. For this, we employ the VAE encoder to generate additional training data points that are regularized by the MMD term, which helps to enhance the model's performance and prevent overfitting. The VAE is utilized to generate $n_m$ new samples in the latent space $\hat{\z}_j$ as follows (described in Algorithm \eqref{alg:PCE_training}). First, we sample $n_m$ points from the encoder's prior  distribution $\mathcal{N} \left( \mat{0}, \mat{I} \right)$, we then use the decoder $D_{\theta}(\cdot)$, followed by the encoder $E_{\vphi}(\cdot)$ to obtain samples $\hat{\z}_j, \, j=1, \dots, n_m$ (in most of our experiments, we consider $n_m\sim1000$ samples). Using these samples, we train the model with MMD regularizer to ensure the responses capture the underlying distribution, as opposed to relying on point-wise estimates from the samples. In addition, since the VAE framework encourages the posterior distribution to closely resemble the standard normal prior, the newly generated samples align with the input distribution.

Let us denote
\begin{equation*}
\hat{y}_j = \frac{1}{n_m} P_{\ell} (\hat{\z}_j) = \sum_{k=1}^{\ell_p} c_{k} \varphi_{k} (\hat{\z}_j)  \, .
\end{equation*}
Then, the $c_{k}$'s coefficients can be estimated by minimizing the loss function
{\small
\begin{align} \label{loss_PCE_MMD}
{\cal L}_{\sigma, \lambda} & = \frac{1}{n_s}\sum_{i=1}^n \left(  \tilde{y}_i - y_i \right)^2 + \lambda \Bigg( \frac{1}{n^2} \sum_{i,j=1}^n \mathcal{K}\left (y_i,y_j \right) \\ \nonumber &   -\frac{2}{n n_m} \sum_{i=1}^n \sum_{j=1}^{n_m} \mathcal{K}\left( y_i,\hat{y}_j \right) + \frac{1}{n_m^2}  \sum_{i,j=1}^{n_m} \mathcal{K}\left( \hat{y}_i, \hat{y}_j \right)\Bigg)^{1/2}\, ,
\end{align}}%
where $\mathcal{K}(y,y')= \exp{(-(y - y')^2/ 2 \sigma^2)}$ is the Gaussian kernel, and $\sigma$ and $\lambda$ are hyperparameters to be tuned. Choosing the Gaussian kernel achieves moment matching of all orders. Moreover,  the square root $\sqrt{{\cal L}_{{\text{MMD}}^2} }$ in Eq. \eqref{loss_PCE_MMD} captures the dissimilarities between distributions better for small values, see~\cite{li2015generative} for details.  

After the dimensionality reduction step using VAE (or $\beta$-VAE),  we employ a bilevel alternating minimization procedure to jointly learn the PCE coefficients $c_{k}$ and the hyperparameters $[\lambda,\sigma]$ related to the MMD regularizer; see Appendix~\ref{app:BAM} for details.
The PCE-Net model with MMD regularization is depicted in Figure \ref{graph_model_PCENet}, and the training procedure is detailed in Algorithm~\ref{alg:PCE_training}.
\begin{algorithm}[!tb]
\caption{\label{alg:PCE_training} PCE-Net with MMD regularization}
\begin{algorithmic}
\STATE \textbf{Input}: $(\x_{1},y_{1}),\dots,(\x_{n},y_{n}) \in \mathbb{R}^{D}\times\mathbb{R}$, latent space dimension $d$, PCE degree $\ell$, regularization parameter set $\mathcal{S}_\lambda$, hyperparameter set $\mathcal{S}_\sigma$.
\STATE \textbf{Output}: PCE output $\tilde{y}_1,\dots,\tilde{y}_n$.
\STATE \textbf{1.} Split the data into train, validation, and test sets
\STATE \textbf{2.} Train a VAE network using the training inputs:\\
 $\vmu(\x_i), \vsigma^2(\x_i) \gets E_{\vphi} (\x_i): \forall i=1,\dots,n$.
\STATE \textbf{3.} Set $p_{\vphi} (\z| \x_i) \leftarrow \mathcal{N} \left( \vmu(\x_i), \vsigma^2(\x_i) \right)$ and sample $\z_i^{(j)} \sim p_{\vphi} (\z| \x_i)$ for $j=1,\dots,n_s$
\STATE \textbf{4.} Generate samples for MMD:
\STATE \quad \textbf{(a)} Sample $\z_j \sim \mathcal{N} \left( \mat{0}, \mat{I} \right)$ for $j=1,\dots, n_m$
\STATE \quad \textbf{(b)} $  \vmu(\z_i), \vsigma^2(\z_i) \gets D_{\theta}(\z_j) $
\STATE \quad \textbf{(c)} Sample $ \tilde{\x}_j \sim \mathcal{N} \left( \vmu(\z_j), \vsigma^2(\z_j) \right)$
\STATE \quad \textbf{(d)} $ \vmu(\tilde{\x}_j), \vsigma^2(\tilde{\x}_j) \gets E_{\vphi}(\tilde{\x}_j)$
\STATE \quad \textbf{(e)} Sample $ \hat{\z}_j \sim \mathcal{N} \left( \vmu(\tilde{\x}_j),  \vsigma^2(\tilde{\x}_j) \right) $
\STATE \textbf{5.} Use $\z_i^{(j)}$ and $\hat{\z}_j$ to jointly optimize $\lambda \in \mathcal{S}_\lambda,\sigma \in \mathcal{S}_\sigma$ and PCE coefficients by bilevel alternating minimization
\STATE \textbf{6.}  $\forall i=1,\dots,n: \ \tilde{y}_i \gets \frac{1}{n_s} \sum_{j=1}^{n_s} P_{\ell} (\z_i^{(j)})   $
\RETURN $\tilde{y}_1,\dots,\tilde{y}_n$
\end{algorithmic}
\end{algorithm} 




\paragraph{High Order Moments of the Responses}
Since we use PCE as the surrogate model, in addition to being able to compute point-wise responses at a given data point, we can also explore the global behavior of the model through the high-order moments of that response. The $k$th moment of the PCE response $\tilde{y}_i$ is given by
\begin{equation} \label{moments}
    m_k(\tilde{y}) = \int_{\mathbb{R}^d} (P_{\ell}(\z))^k f_{\matZ}(\z) d\z \, ,
\end{equation}
where $f_{\matZ}(\z)$ is the joint standard normal probability density. The above integral can be computed using the orthogonality of the polynomial basis. For example, the mean and variance are given by $ \mu_Y = c_1, \, \sigma_Y^2 = \sum_{i=2}^{\ell_p} c_i^2 \, .$ 
Moreover, when the MMD regularizer is used, it promotes  these moments of the PCE response to match the moments of the true outputs. Using the results in \cite{chen2021analysis}, we can argue that by matching the moments we can ensure that the distribution of the response $\tilde{y}$ is close to the distribution of the output $y$.
In particular, Proposition 1 in~\cite{chen2021analysis} says, for any two probability density measures $\veta$ and $\vnu$ that have the same moments up to degree $k$, and are constant on some interval $[a,b]$,  the following holds: 
$$\int \left| \veta(\z) - \vnu(\z) \right| d\z < 12(b-a)k^{-1}.$$
Indeed, MMD aims to minimize the (Wasserstein) distance between the true distribution of the outputs and the distribution of the surrogate model, by matching moments. Therefore, the above distance between the distributions (of $\tilde{y}$ and ${y}$) is likely to be small and bounded. 

\paragraph{Properties of PCE-Net}
Here we list a few properties and advantages of \texttt{PCE-Net} over other existing UQ methods:
\begin{itemize}
    \item \texttt{PCE-Net} (in contrast to other DRSM models) computes the distributions of data in the latent space (using VAE), and we do not need to approximate the distribution of the input data (e.g. with kernel density estimation). Hence, it is computationally less expensive, since we only need ${(\ell+m)!}/{\ell!m!}$ coefficients in contrast to ${(\ell+d)!}/{\ell!d!}$ when using PCE without VAE.
    \item \texttt{PCE-Net} yields point-wise estimates by using an $L_2$ term in the loss function but also captures the global behavior of the outputs (distribution of output) via the MMD regularizer which aims to match moments. 
    
    \item The use of VAE  is advantageous since we are able to generate additional samples for training. Thus, we can also capture system dynamics efficiently in the latent space. 
    
     \item We can compute the moments of the global output variable directly from PCE  using the coefficients, e.g., the first moment (mean) is the first coefficient of PCE, variance (second moment) is the sum of the square of the coefficients, and so on. However, in order to compute the conditional moments (Eq. \eqref{moments}), e.g., by Monte-Carlo of the moments' integral, other methods will need  to approximate the (high-dimensional) data distribution, whereas \texttt{PCE-Net}  only requires sampling from the latent space. 

\end{itemize}

\section{Numerical Results}\label{sec:expts}
In this section, we present numerical results illustrating the performance of \texttt{PCE-Net} on various datasets from different applications. We begin by considering the double pendulum problem and present a UQ study for this problem using \texttt{PCE-Net}. We then present results for three datasets that are widely used in the context of machine learning, and two datasets related to solving PDEs. We also present results for robust learning using \texttt{PCE-Net}. For the implementation of VAE, we used the Pyro package~\cite{bingham2019pyro}, and for the PCE implementation, we used the Chaospy package~\cite{feinberg2015chaospy}.

 \subsection{Quantifying uncertainty in experimental data} \label{uq_in_experiments}
We begin with the double pendulum problem discussed in the introduction. The goal is to quantify the uncertainty of the dynamical system directly from the observed data. Motivated by the work in~\cite{chen2022automated}, we use \texttt{PCE-Net} to encapsulate a relationship between current and future states of a double pendulum and extract the hidden state variables. We assume that the equations of motion of the system described below in Eq.~\eqref{eq:double_pendulum_equation} are not known, and we aim to learn a latent distribution that describes the dynamics of the system represented by the images of the pendulum at consecutive times. By using \texttt{PCE-Net}, we can capture distributions of different states due to varying conditions. In other words, we can learn low-dimensional dynamics under uncertainty. In~\cite{chen2022automated}, a geometrical learning approach is proposed for determining how many state variables an observed system is likely to have. This can also be applied here. However, for the sake of demonstration, we set the latent space size to be equal to the stochastic dimension of the problem.

\begin{figure}[t!]
\centering{}
\includegraphics[clip,scale=0.3]{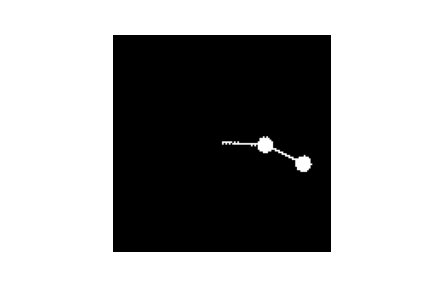} \includegraphics[clip,scale=0.3]{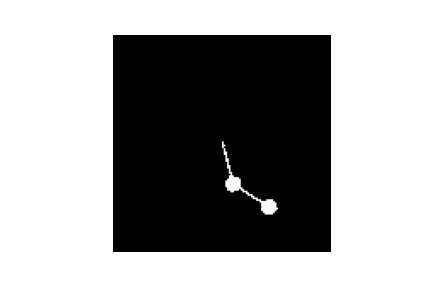}
\includegraphics[clip,scale=0.3]{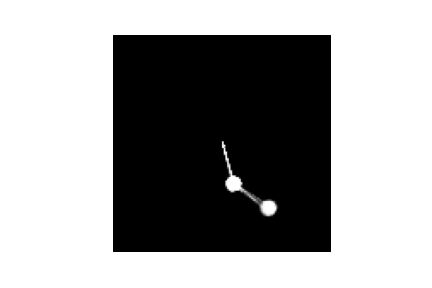}

\includegraphics[clip,scale=0.3]{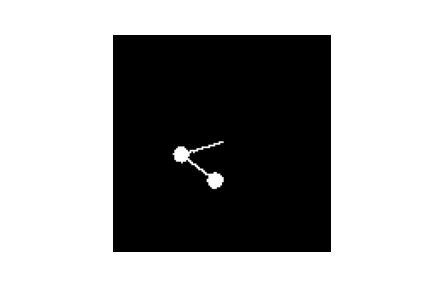} \includegraphics[clip,scale=0.3]{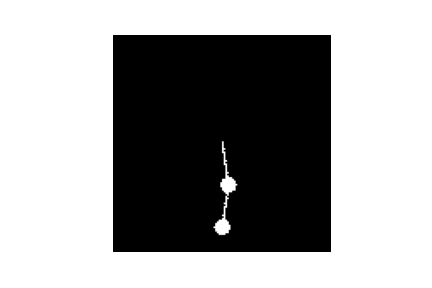}
\includegraphics[clip,scale=0.3]{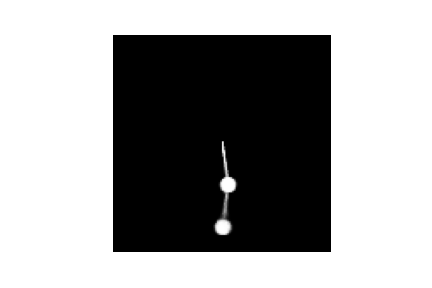}

\includegraphics[clip,scale=0.3]{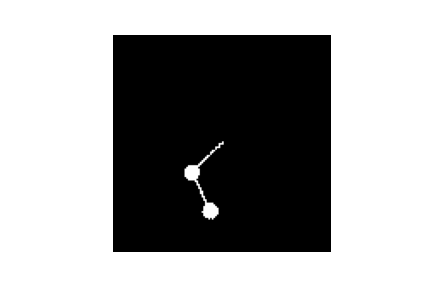} \includegraphics[clip,scale=0.3]{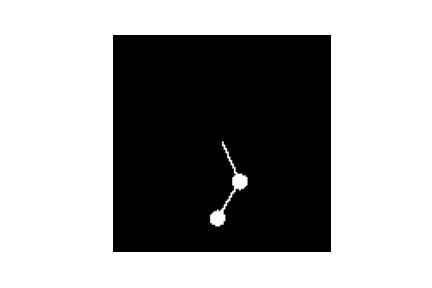}
\includegraphics[clip,scale=0.3]{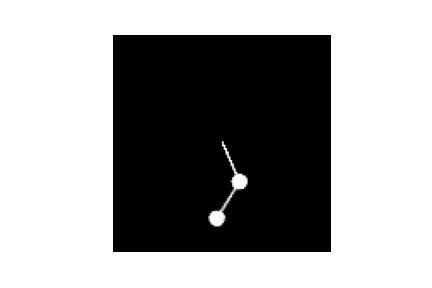}
\caption{Example images of a double pendulum at the initial random states (left), after time $20 \Delta t$ (middle), and the VAE reconstruction of the output obtained (right).}\label{fig:double_pendulum_images}
\end{figure}

\begin{figure}[t!]
\centering{}
\includegraphics[clip,scale=0.42]{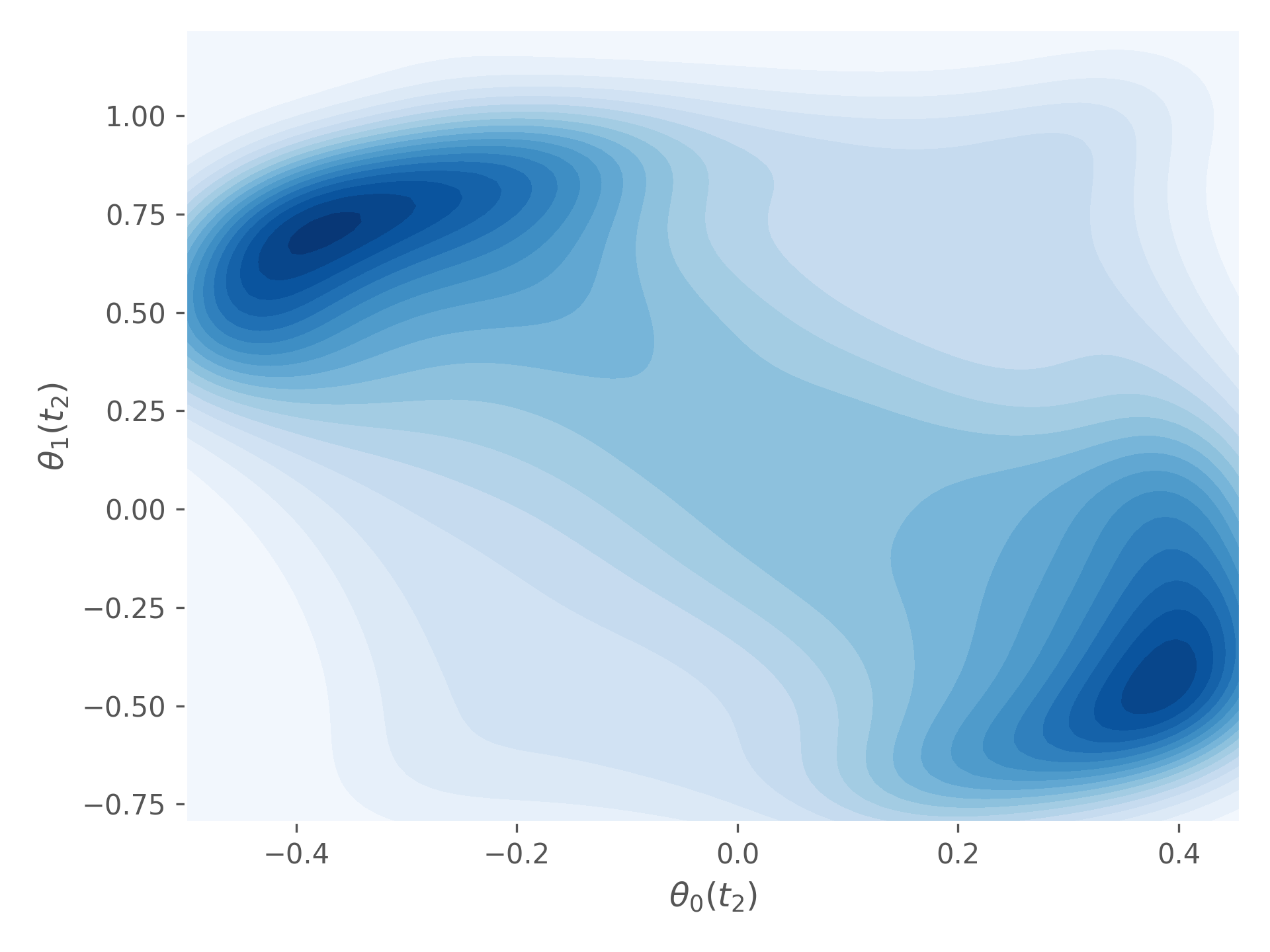} \includegraphics[clip,scale=0.42]{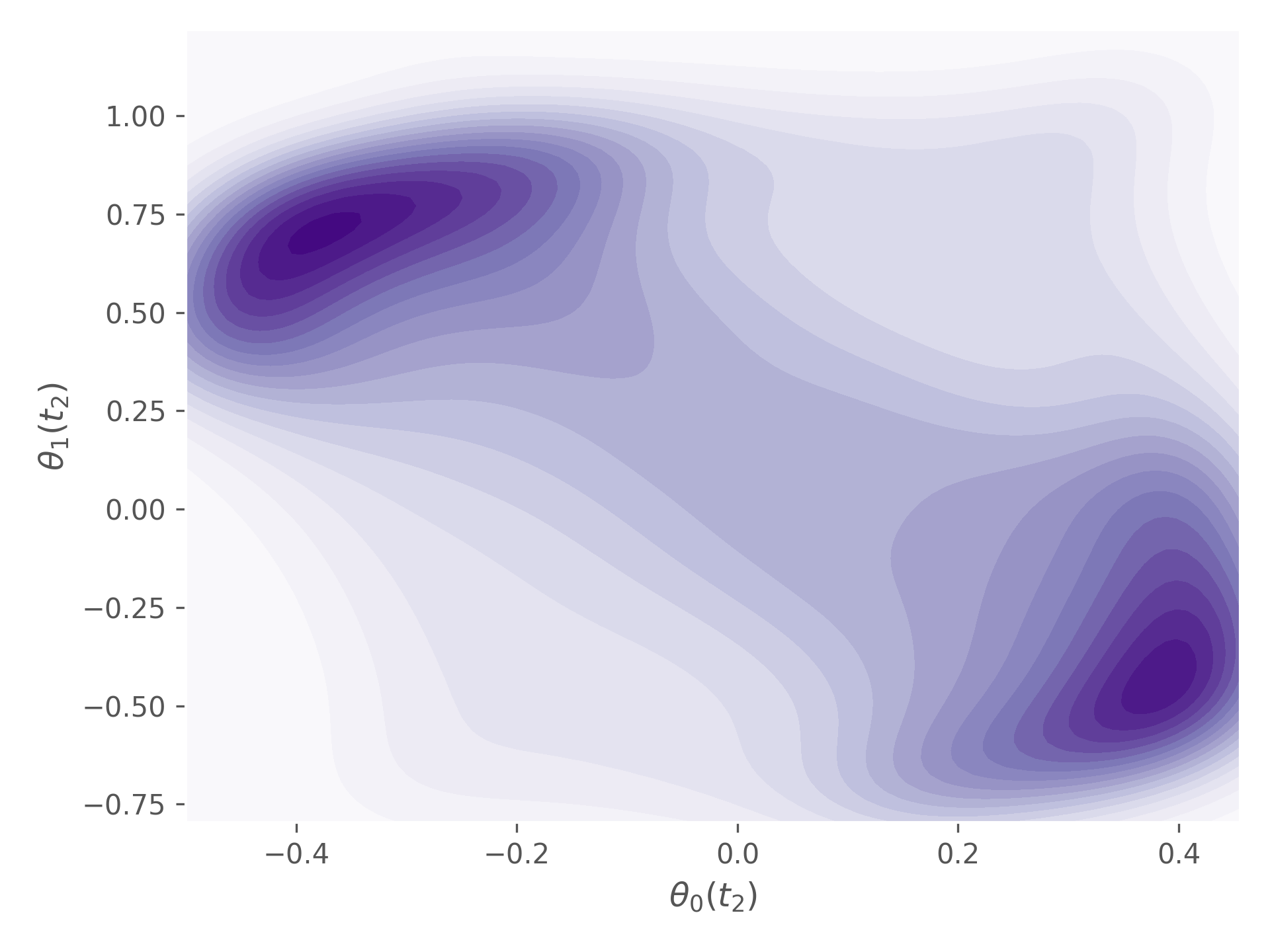}
\caption{The PDF contours for the two angles computed using 10000 Monte Carlo simulations (left) and the PCE surrogate (right). The PCE coefficients were calculated through regression using 49 Gaussian quadrature nodes (polynomial order 6).}\label{fig:PDF_contours}
\end{figure}

We define the pendulum rod lengths, and top and bottom bob masses as $l_1=0.5$, $l_2=0.5$, $m_1=2$ and $m_2=2$, respectively. The two degrees of freedom are the angle between the top rod and the $y$-axis, $\theta_1$, and the angle of the second pendulum, $\theta_2$. In Cartesian coordinates, we compute the positions of the bobs as follows:
\begin{equation}
    \begin{matrix}
    x_1 &=& l_1 \mathrm{sin} \theta_1& \dot{x}_1&=&l_1 \dot{\theta}_1 \mathrm{cos} \theta_1,\\ 
    y_1 &=& -l_1 \mathrm{cos} \theta_1 & \dot{y}_1&=&l_1 \dot{\theta}_1 \mathrm{sin} \theta_1,\\ 
    x_2 &=& l_1 \mathrm{sin} \theta_1 + l_2 \mathrm{sin} \theta_2& \dot{x}_2 &=& l_1 \dot{\theta_1} \mathrm{cos}\theta_1  + l_2 \dot{\theta_2}\mathrm{cos} \theta_2,\\ 
    y_2 &=& -l_1 \mathrm{cos} \theta_1 - l_2 \mathrm{cos} \theta_2 & \dot{y_2} &=& l_1 \dot{\theta_1} \mathrm{sin} \theta_1 + l_2 \dot{\theta_2} \mathrm{sin} \theta_2.
    \end{matrix}
\end{equation}
The equations of motion for the double pendulum are often written using the Lagrangian formulation of mechanics and solved numerically. Denoting gravitational acceleration as $g$, we can then define the kinetic energy, $T=\frac{1}{2} m_1 v_1^2 + \frac{1}{2} m_2 v_2^2$, and potential energy, $V=m_1 g y_1 + m_2 g y_2$. The Lagrangian is $L=T-V$ and the equation of motion (the Euler-Lagrange equation) is
\begin{equation}
\frac{d}{dt}\left (  \frac{\partial L}{\partial \dot{q}_i}\right ) - \frac{\partial L}{\partial q_i} 0 , \quad \mathrm{where} \quad q_i \in \left \{  \theta_1, \theta_2 \right \}.
\end{equation}\label{eq:double_pendulum_equation}
The dynamics of the double pendulum can be described by four variables, the two angles and the corresponding angular velocities (momenta). In the study, we generate images of a swinging double pendulum for varying initial angles. We assume initial angles are uniformly distributed over $\mathcal{U}(-\pi/2, \pi/2)$. The positions at $t_1$=0 are sampled from a joint distribution. We then solve the set of nonlinear differential equations describing the motion of the system, and output images of the double pendulum at the time $t_1=0$  and  $t_2 = 20 \Delta t$, where $\Delta t=0.02$. Example input-output pairs are shown in the first two columns of Fig.~\ref{fig:double_pendulum_images}. The equations are solved using a Python package, \texttt{ODEINT} in SciPy \cite{2020SciPy-NMeth}.

First, we consider the two-dimensional problem in Eq.~\eqref{eq:double_pendulum_equation} directly, and compare the PCE surrogate modeling approach against (traditional and more expensive) Monte Carlo simulations. Figure~\ref{fig:PDF_contours} shows the PDF contours obtained for the two output angles using 10000 Monte Carlo (MC) simulations  on the left, and the result obtained by the PCE surrogate model (degree = 6) on the right. The means and the standard deviations of the distributions for the two angles obtained using the MC approach were $\mu_{MC}(\theta_1) = 0.0172,\mu_{MC}(\theta_2) = 0.159; \sigma_{MC}(\theta_1) = 0.283, \sigma_{MC}(\theta_2) = 0.505$, respectively, and by PCE approach were $\mu_{PCE}(\theta_1) = 0.0168,\mu_{PCE}(\theta_2) = 0.159; \sigma_{PCE}(\theta_1) = 0.284, \sigma_{PCE}(\theta_2) = 0.506$.  We observe that the two contours match fairly well, even though the PCE approach is significantly less expensive computationally compared to MC simulations. However, we can use such a PCE approach only when we have small number of input variables, and their distributions (PDFs) are known. In our original double-pendulum problem,  we only have access to the initial positions of the pendulum in the form of images of size $128\times128$ pixels, therefore, UQ methods such as the (standard) PCE would be infeasible, in order to construct a surrogate that maps the input PDFs to the PDFs of output angles. Moreover, in this case, we assume that the relevant set of input variables is unknown. Hence, we deploy \texttt{PCE-Net} for this problem.

\begin{figure}[!t]
\centering{}
\includegraphics[clip,scale=0.42]{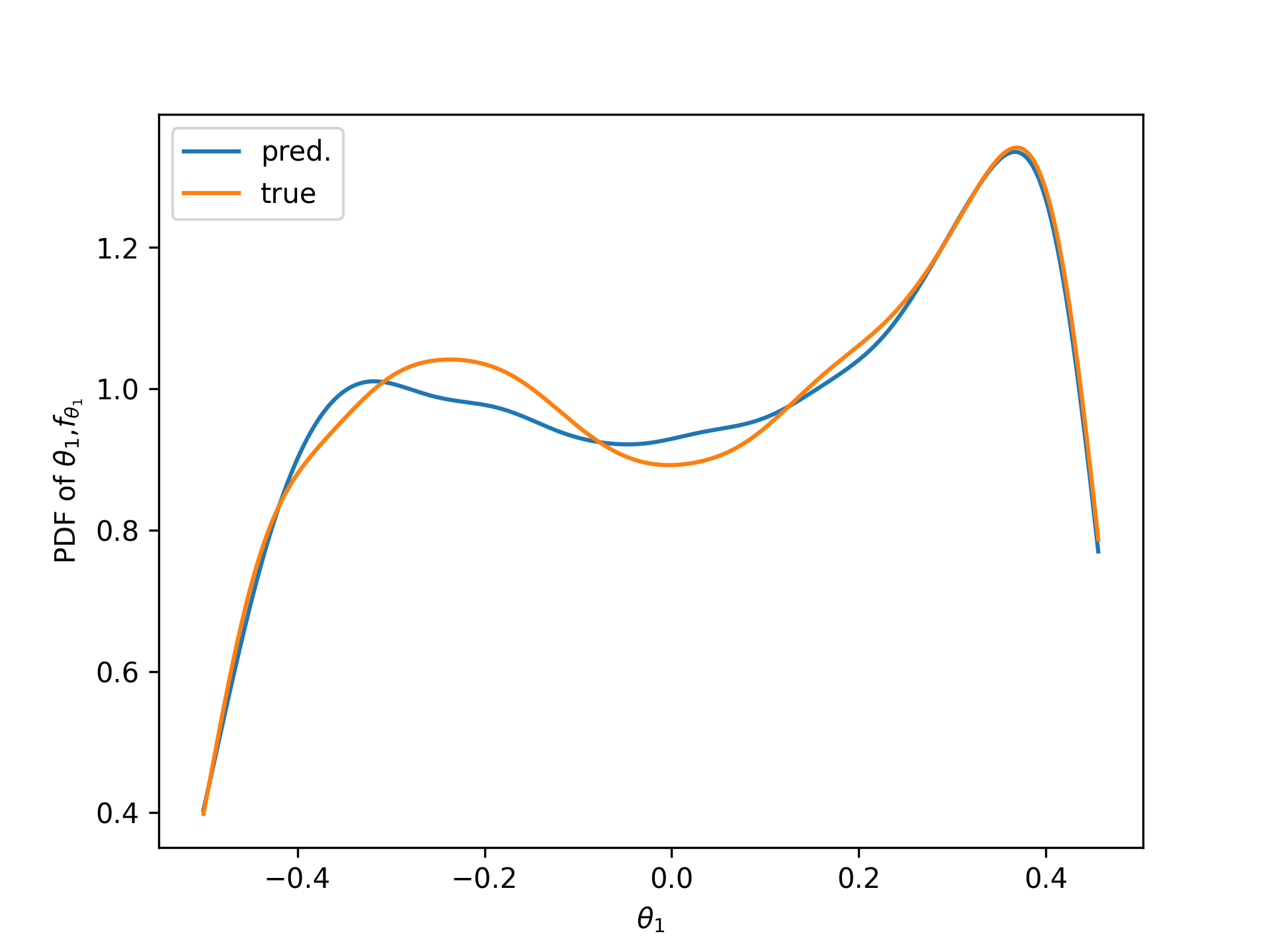} \includegraphics[clip,scale=0.42]{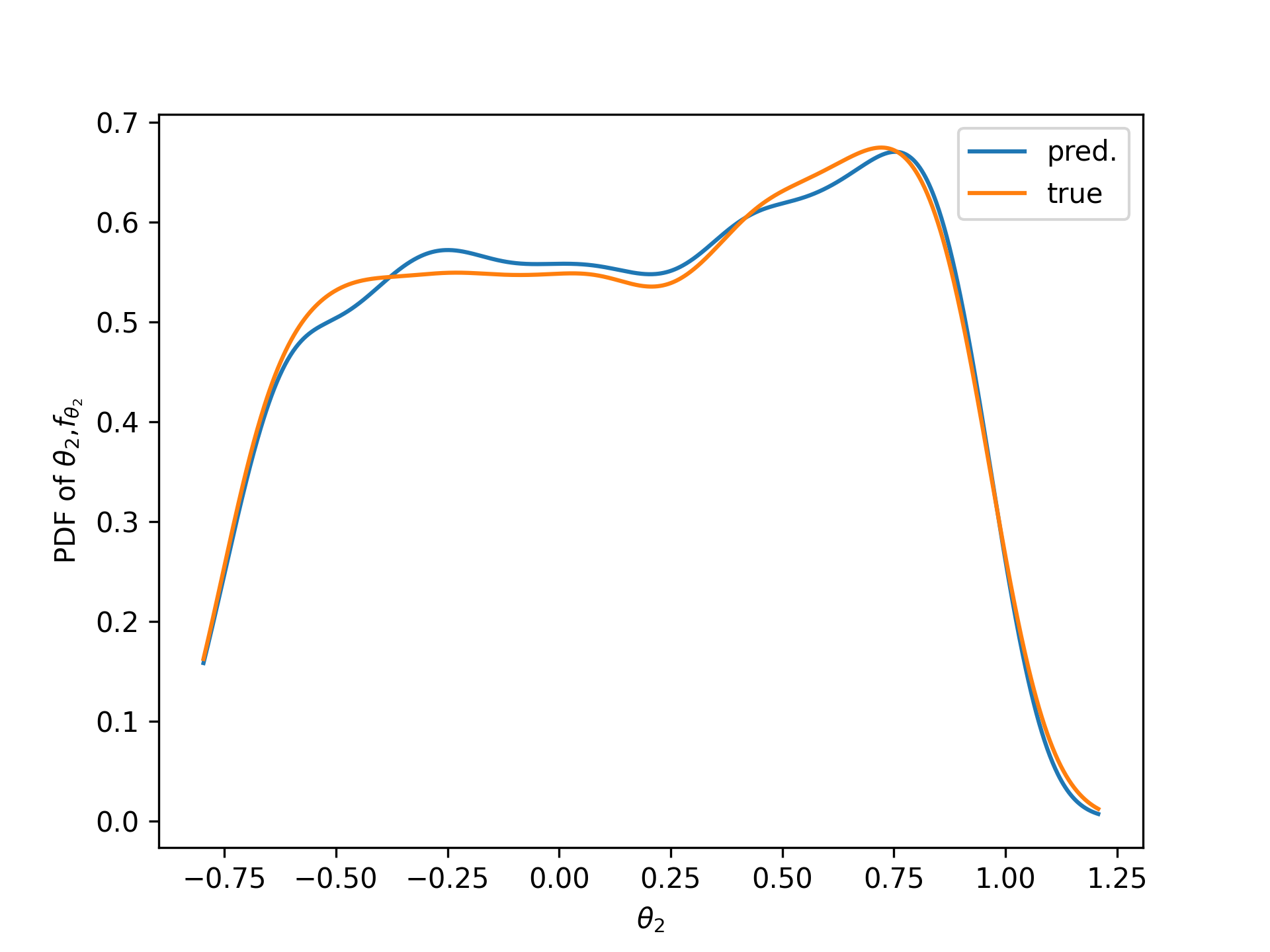}
\caption{Predicted PDFs of two output angles at $t_2$ using PCE-Net (the PCE surrogate obtained from VAE latent distribution).}\label{fig:double_pendulum_output_pdfs}
\end{figure}

We use the images to train a VAE, where inputs are the $128\times128$ snapshots of the double pendulum at $t_1$ and outputs are the images at $t_2 = t_1+20\Delta t$. We use a $\beta$-VAE that introduces an additional hyperparameter $\beta$, which controls the trade-off between the reconstruction error and the KL-divergence term in the loss function, as discussed in Section~\ref{par:disentanglement}. In our approach, we enforce the latent distribution to resemble a predefined prior $\mathcal{N}(0, 1)$, making the surrogate construction more effective (hence we use $\beta$-VAE). Example reconstructed images obtained from the $\beta$-VAE decoder for the corresponding inputs are given in the right column of Figure \ref{fig:double_pendulum_images}. We then construct the PCE surrogate by sampling from the learned joint two-dimensional distribution, which inherently captures the dynamics of the system (characterized by changing values of the angles).

We test the system using 6500 images. We construct the surrogate based on additional numerical descriptors: 100 values of output angles at time $t_2$. So, the Hermite PCE coefficients are approximated by passing the corresponding 100 input images at $t_1$ through the $\beta$-VAE, sampling from the resulting latent distributions and matching the outputs with the angles at $t_2$. We set the value of $\beta=20$, the latent dimension is 2, and the polynomial degree is 6. We use the square loss function for coefficient learning. Figure~\ref{fig:double_pendulum_output_pdfs} plots the output PDFs for the two angles at time $t_2$ computed by our method, as well as the true PDFs. We note that \texttt{PCE-Net} approximates the output PDFs of given angles extremely well. A few realizations of the dynamics of the parameters ($\theta_1$ and $\theta_2$) over two different time-scales are given in Appendix~\ref{app:dp}.

\subsection{Machine Learning Datasets} \label{ML_datasets}
 We next demonstrate the performance of \texttt{PCE-Net} on three datasets that frequently appear in machine learning applications, and are used for analyzing supervised learning methods. These datasets are of moderate to high-dimensions, and traditional UQ or surrogate modelling methods do not scale to such problems. Thus, the use of more sophisticated methods such as \texttt{PCE-Net} is necessary.

\textbf{Error metrics: \label{errors_metrics}} To analyze the distribution learned by the model and compare the responses histograms, we consider two error metrics, namely, the Wasserstein distance (optimal transport)~\cite{villani2021topics} given by:
\begin{align}\label{wassersteain}{\displaystyle W_{p}(X,Y)=\left(\sum _{i=1}^{n}\|X_{(i)}-Y_{(i)}\|^{p}\right)^{1/p}},\end{align}
where $X_{(i)}$'s and $Y_{(i)}$'s are the order statistics of $X$ and $Y$, respectively, and the Mahalanobis distance~\cite{de2000mahalanobis} that measures the distance of a point $x$ from a probability distribution with mean $\vmu$ and covariance matrix $\matC$ is given by:
\begin{equation} \label{mahalanobis}
    M(\x, \vmu) = \sqrt{(\x - \vmu)^T \matC^{-1}(\x - \vmu)} \, .
\end{equation}
The Wasserstein-1 distance can also be used to quantify point-wise estimation errors for regression problems, which we  utilize to quantify errors for a neural network (MLP) method in the numerical experiments below.
In the \texttt{PCE-Net} model, we compare the true responses $y_i$ and the model's learned responses $\tilde{y}_i$, by considering Eq. \eqref{wassersteain} with $p=1$, and by  summing over all the Mahalanobis distances $M(\mathrm{dist}(\tilde{y}_i),y_i)$ with the mean and variance of each $\tilde{y}_i$:
 {\small \begin{equation*}
    \mu_{\tilde{y}_i} = \frac{1}{n_s} \sum_{j=1}^{n_s} P_{\ell} (\z_i^{(j)}) \, ,\, \matC_{\tilde{y}_i} = \frac{1}{n_s} \sum_{j=1}^{n_s} \left( P_{\ell} (\z_i^{(j)}) - \mu_{\tilde{y}_i} \right) ^ 2 .
\end{equation*} }
Since these metrics account for the distance between distributions, they are popularly used in many UQ tasks and applications~\cite{candelieri2022use,bi2017uncertainty,vishwakarma2021metrics}.

\begin{figure*}[!tb]
\centering{}%
\includegraphics[scale=0.3]{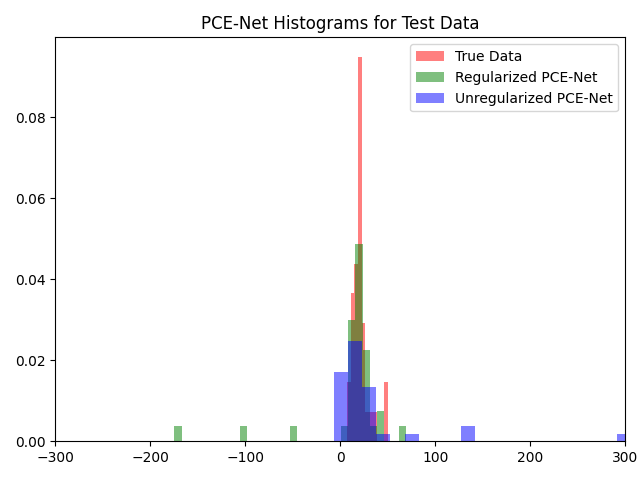}
\includegraphics[scale=0.3]
{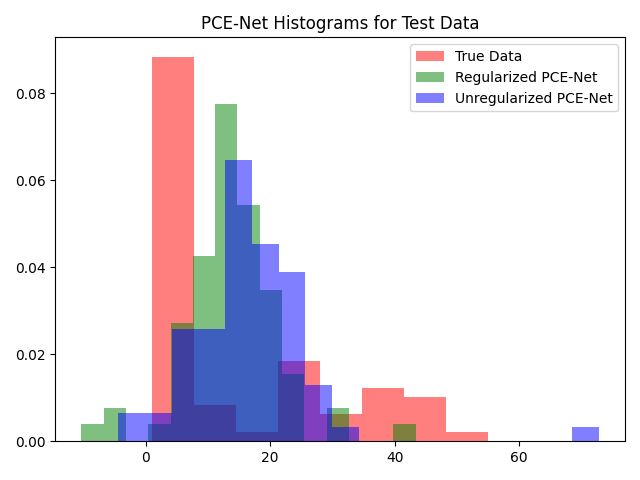}
\includegraphics[scale=0.3]{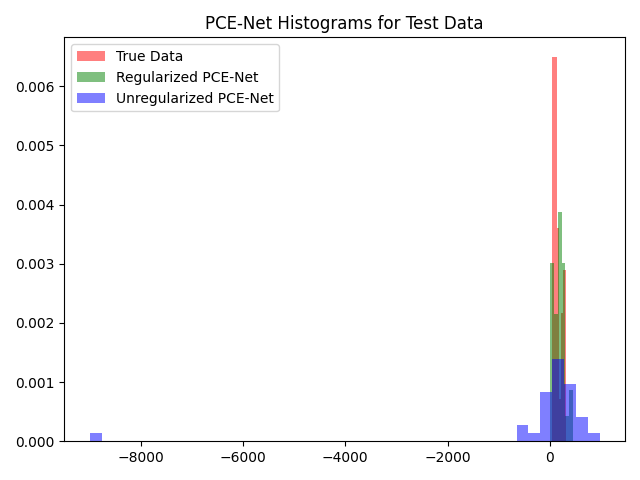}
\caption{\label{Real-fig}Histogram plots of the responses test data. Left to Right: Boston housing, Parkinson Speech, Diabetes.}
\vskip -0.1in
\end{figure*}
{\bf Implementation details:}
In the following experiments, we used Algorithm \ref{alg:PCE_training} for training \texttt{PCE-Net} with the following components. The VAE consisted of a symmetric encoder and decoder, each with one to three hidden layers and with the activation functions softplus and sigmoid, respectively. The training was performed with an Adam optimizer. For the PCE stage, since ${\cal L}_{\sigma, \lambda}$ in Eq. \eqref{loss_PCE_MMD} is linearly dependent on $\lambda$, we performed grid search for $\lambda$ with Wasserstein-1 loss in Eq. \eqref{wassersteain} over the set $\mathcal{S}_\lambda=\{10^{-7}, \dots, 10^{-1}, 1,10\}$. Next, we learned the PCE coefficients and the hyperparameter $\sigma$ using  a bi-level alternating  minimization (BAM) formulation. 
 The set of $\sigma$ values that were considered is $\mathcal{S}_\sigma=[0.1,5]$ (which captures the median proximity heuristic for all the datasets, see \cite{gretton2006kernel, muandet2017kernel}).  The results for the three ML datasets  are presented in Figure~\ref{Real-fig} and Table~\ref{errors-table}. 
In Table~\ref{errors-table} and \ref{robust-table}, we compare the results of PCENet to two neural network (NN) methods, namely (a) a fully connected 2-layer neural network/Multilayer perceptron (MLP) with ReLU activation function, and (b) an alternate DRSM approach, VAE-MLP, where the variational autoencoder (VAE) is used for dimensionality reduction (similar to PCENet), and a fully connected 2-layer neural network (MLP) is used as a surrogate model to learn a mapping from the latent space to the outputs. The MLP method is a supervised regression approach and yields point-wise estimates for the test data. Therefore, we only report Wasserstein-1 error for this method in the tables. For VAE-MLP, we used a Kernel Density Estimation (KDE) approach with a Gaussian kernel to obtain the output distributions for the test datasets, and then compute both the Wasserstein and Mahalanobis errors based on these distributions.


{\bf Boston Housing:}
This dataset concerns the housing values in the suburbs of Boston, 1970. It was published in \cite{harrison1978hedonic}, and has been featured in many machine learning articles that address regression problems, including analysis with different supervised learning methods in \cite{quinlan1993combining}.
The data consists of 506 samples with 13 features. The features were encoded with a VAE (6 neurons in the hidden layer) into a 3-dimensional latent space. The learning rate (LR) was $10^{-3}$ and the number of epochs (\#epochs) was $1400$. For the PCE stage, we chose  the degree to be 11. The  parameter values we obtained from BAM were $\lambda=0.1$ and $\sigma=0.2342$. The result for this data is reported in Figure \ref{Real-fig} (left). 


{\bf Parkinson Speech:}
The Parkinson dataset consists of 20 People with Parkinson's and 20 healthy people. From all the patients, 26 types of sound recordings (voice samples including sustained vowels, numbers, words, and short sentences) were taken and used as the features (1040 samples overall). The UPDRS (Unified Parkinson Disease Rating Scale) score of each patient is the output that is determined by an expert physician. This dataset was collected and studied in~\cite{sakar2013collection}. We encoded the features using the VAE (13 neurons in the first hidden layer and 7 neurons in the second hidden layer) to a 4-dimensional latent space. The LR was $0.001$ and \#epochs was $3700$. For the PCE stage, degree of 6 was chosen. The values we obtained for parameters $\lambda=10$ and $\sigma=0.1354$. Histogram results are reported in Figure \ref{Real-fig} (center).


{\bf Diabetes Dataset:}
The diabetes dataset consists of 442 diabetes patients, with a baseline of 10 health features
as well as the response of interest, a quantitative measure of disease progression one year after baseline. This dataset was analyzed in \cite{efron2004least}. We embedded the input features into a 3-dimensional latent space using the  VAE (6 neurons in the hidden layer). LR was $10^{-3}$ and \#epochs was $200$. The PCE degree was 11. The hyperparameter values obtained were $\lambda=0.1$ and $\sigma=12.2092$, see Figure \ref{Real-fig} (right) for the histogram results.



In the histogram plots, we note that, in all three cases, (a) the PCENet estimates (distributions) are close to the true point-wise values, and (b) PCENet estimates have  Gaussian type distributions around the true test points, i.e., we obtain distributions rather than point-wise estimates. In the case of Parkinson dataset, the true data histogram is spread-out, and both the PCENet approaches (unregularized and MMD regularized) attempt to fit Gaussian distributions centered close to the true mean of the test data.

\begin{figure*}[!tb]
\centering{}%
\includegraphics[clip,scale=0.4]{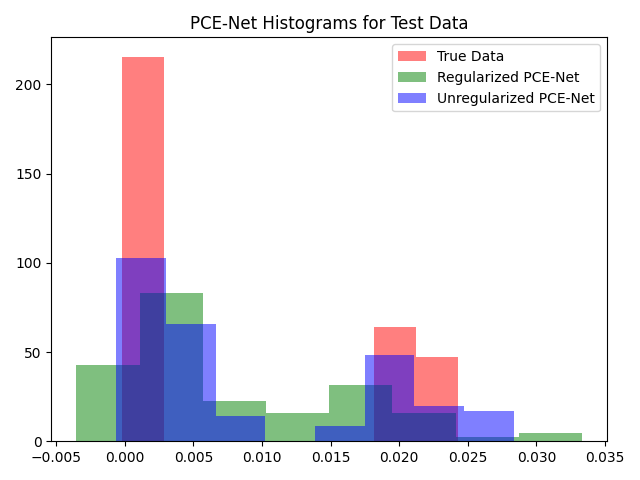} \includegraphics[clip,scale=0.4]{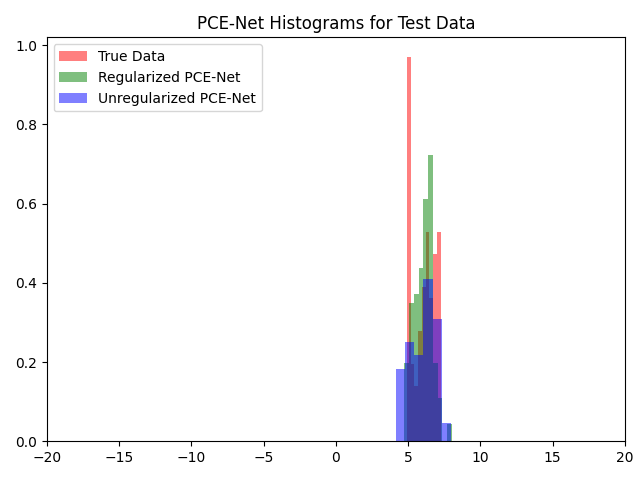}
\includegraphics[clip,scale=0.48]{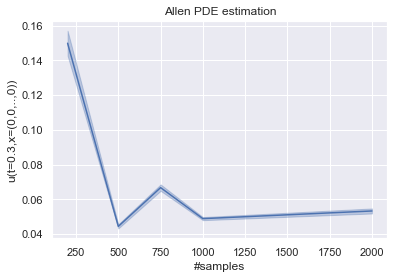} 
\includegraphics[clip,scale=0.48]{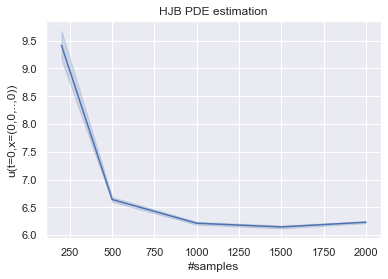} 
\caption{\label{PDE-fig} PDE Results: (Top) Histogram plots of the responses of test data for  Allen-Cahn on the left and Hamilton-Jacobi-Bellman on the right. (Bottom) PCENet estimations (mean and standard deviation) as a function of number of samples in the latent space, for Allen-Cahn equation $u(t=0.3, x=(0, \ldots , 0))$ on the left and for  HJB equation $u(t=0, x=(0, \ldots , 0))$ on the right. }
\vskip -0.1in
\end{figure*}
\subsection{Learning High-dimensional Differential Equations} \label{PDEs-datasets}
Next, we demonstrate how the proposed \texttt{PCE-Net} approach performs on problems that aim to learn solutions of high-dimensional partial differential equations (PDEs). We consider two PDEs, namely Allen-Cahn Equation and Hamilton-Jacobi-Bellman (HJB) Equation. Both PDEs that are considered here appeared in~\cite{han2018solving} as examples for solving high-dimensional PDEs using deep learning. The objective of \texttt{PCE-Net} is to learn the input-output relation defined by the PDEs, using which we can predict the PDE solution for a given input state. The results for these datasets are presented in Figure~\ref{PDE-fig} and Table~\ref{errors-table}.

{\bf Allen-Cahn Equation:} The Allen-Cahn equation is a reaction-diffusion equation, typically used as a prototype for modeling phase separation and order-disorder transition, see~\cite{emmerich2003diffuse,han2018solving} for details. 
We considered the following Allen-Cahn equation for $t \in [0,0.3]$ and $\x \in \mathbb{R}^{100}$:
\begin{align*}
    \frac{\partial u}{\partial t} (t,\x) &= \Delta u(t,\x) + u(t,\x) - u^3(t,\x), \\
    u(t=0,\x) &= \frac{1}{2+0.4 \Vert \x \Vert^2}.
\end{align*}

 We sampled $\x_1,\dots,\x_{1385}$ from different normal distributions, and used the method in~\cite{han2018solving} to obtain the solution $u(t=0.3,\x_1),\dots,u(t=0.3,\x_{1385})$ for the equation. The goal of \texttt{PCE-Net} is to learn the input-output relation $(\x_1,u(t=0.3,\x_1)),\dots,(\x_n,u(t=0.3,\x_{1385}))$. The latent dimension chosen was 6,  the VAE had 50 neurons in the first hidden layer, 25 neurons in the second hidden layer, and 12 neurons in the third hidden layer. LR was $10^{-3}$ and \#epochs was $3700$. We chose  the PCE degree to be 5. The hyperparameter values obtained were $\lambda=10$ and $\sigma=1.007$. Results are reported in Figure \ref{PDE-fig} (left). First, we plot the histogram of the PCENet responses for the test data (top left), similar to the results in Figure~\ref{Real-fig}. We note that the true data histogram has spikes at two regions and PCENet captures these two regions extremely well. Next, we utilize the input-output relation learned by PCENet to predict the PDE solution for a given input state. For the Allen-Cahn equation, we consider the input state $x=(0, \ldots , 0)$ and estimate $u(t=0.3, x=(0, \ldots , 0))$ using PCENet. In Figure~\ref{PDE-fig} (bottom left) we report the PCENet (mean and standard deviation) estimations as a function of number of samples in the latent space. As reported in~\cite{han2018solving}, the approximately computed “exact” solution by
means of the branching diffusion method is $u(t=0.3, x=(0, \ldots , 0)) \approx 0.0528$, and  we note that the PCENet estimates get close to the true solution. 

{\bf Hamilton-Jacobi-Bellman Equation:} The Hamilton-Jacobi-Bellman (HJB) equation appears in many different areas including economics, behavioral science, computer science, and biology. High-dimensional HJB equations are popular in game theory and dynamic resource allocation problems~\cite{han2018solving, powell2007approximate}.
Here, we considered the following Hamilton-Jacobi-Bellman equation for $t \in [0,1]$ and $\x \in \mathbb{R}^{100}$:
\begin{align*}
    &\frac{\partial u}{\partial t} (t,\x) + \Delta u(t,x) - \Vert \nabla u(t,\x) \Vert^2 = 0, \\
   & u(t=1,\x) = \ln\left( \frac{1+\Vert \x \Vert^2}{2} \right).
\end{align*}
The equation minimizes the cost functional through the control process, where the value of the solution $u(t, x)$  at $t = 0$
represents the optimal cost when the state starts from $x$.
We sampled $\x_1,\dots,\x_{1990}$ from different normal distributions, and used the method in \cite{han2018solving} to obtain  $u(t=0,\x_1),\dots,u(t=0,\x_{1990})$ from the equation. The goal of PCE-Net is to learn the input-output relations $(\x_1,u(t=0,\x_1)),\dots,(\x_n,u(t=0,\x_{1990}))$. The 100 features were encoded with a VAE (16 neurons in the hidden layer) to a 6-dimensional latent space.  LR$=10^{-3}$, \#epochs$=4900$, and the PCE degree was 5. The hyperparameter values obtained were $\lambda=10$ and $\sigma=0.4837$. Results are reported in Figure \ref{PDE-fig} (right).
We plot the histogram of the PCENet responses for the test data (top right), which show good agreement with the true data histogram. We again consider the input state of $x=(0, \ldots , 0)$ and estimate the optimal cost $u(t=0, x=(0, \ldots , 0))$ using PCENet. In Figure~\ref{PDE-fig} (bottom right) we report the PCENet (mean and standard deviation) estimations as a function of number of samples in the latent space.

\begin{table}[!tb]
\caption{\label{errors-table} Test errors w.r.t. the two error metrics (Wasserstein and Mahalanobis) for unregularized and regularized PCE-Net, 2 layer MLP, and  VAE-MLP. }
\begin{center}

\begin{tabular}{c|c|c|c|c|c|c|c}
\toprule
& \multicolumn{4}{c|}{Wasserstein} & \multicolumn{3}{c}{Mahalanobis}\\
\cline{2-8}
& \multicolumn{2}{c|}{PCENet} & & & \multicolumn{2}{c|}{PCENet}\\
\cline{2-8}
Datasets & Unreg & Reg & MLP & VAE-MLP & Unreg & Reg  & VAE-MLP  \\
\hline
\hline
Allen-Cahn &  0.0006 & \textbf{0.0005}   &0.0752&0.0494 & 0.2756 & \textbf{0.1649}& 0.1838\\
\hline
HJB & 20.309 & \textbf{0.1815} &6.2093&6.1363& 3.2644 &\textbf{1.2746}&2.3393  \\
\hline
Diabetes & 135.68 & \textbf{9.5943} &144.64&144.77& 0.2284 & \textbf{0.0721}&  11.120\\
\hline
Parkinson & 9.895 & \textbf{9.0144} &13.223&13.306& 1.2241 & \textbf{1.0362}& 2.4902 \\
\hline
Boston & 57.644 & \textbf{19.232} & 20.998 & 20.719 & 0.3012 & \textbf{0.1952}& 1.6081 \\

\bottomrule
\end{tabular}

\end{center}
\end{table}

Next, we report the results w.r.t. the test error metrics \eqref{wassersteain} and \eqref{mahalanobis} in Tables \ref{errors-table} and \ref{robust-table}, where each of the metrics was averaged over five independent trials for each dataset. We compare the results of the two versions of PCENet (unregularized and MMD regularized, respectively) with the two NN approaches, MLP and VAE-MLP. Table \ref{errors-table} demonstrates the errors obtained by the training procedure described in Algorithm \ref{alg:PCE_training}. As expected, these results demonstrate that adding the MMD as a regularization term consistently improves the Wasserstein metric, and in some cases also improves the Mahalanobis metric.  Morevoer, we note that in all cases, PCENet performs better than the two NN approaches with respect to both error metrics, which measure the distances between distributions. This suggests that PCENet performs well in capturing the output distributions and the mapping between I/O distributions. PCENet has several other advantages over the other NN approaches, namely, (a) we can compute output distribution and posterior statistics by sampling points in the latent space and passing them through PCE (we do not need expensive kernel density estimations), (b) we can compute the moments of the global output variable directly from PCE using the coefficients, e.g., the mean (first moment) is the first coefficient of PCE, variance (second moment) is the sum of the square of the coefficients, and so on (higher-order moments), (c) we obtain a functional representation of the input-output relationship, and (d) we can tailor the choice of the polynomial based on the data distribution and can even fit arbitrary distributions without any prior statistical assumptions on the data.

%
%

 \subsection{Robust Learning} \label{robust_learning}
 In this section, we demonstrate the robustness of \texttt{PCE-Net} model. For that purpose, we re-trained the model (only the PCE learning stage is required) for all five datasets, where $5\%$ outliers were added to the training data. The outliers were produced by randomly adding noise with the magnitude of three empirical standard deviations of the dataset. The model parameters ($\lambda$ and $\sigma$) for the datasets were chosen via the BAM approach with the $W_{1}$ metric (see Appendix~\ref{app:RL}).   
Since the Wasserstein metric measures similarity between distributions, it would be better indicative of measuring the robustness. As done before for $W_{1}$, $\lambda$ and $\sigma$ are learned via grid search. In Table \ref{robust-table}, we report the test error metrics for the five datasets under these outlier settings. It can be seen that the MMD regularization term helps in preserving the robustness since it learns and maps distributions as opposed to point-wise estimation. We observe that our model mostly achieves better errors.  
Another way to assess the robustness of outliers in the training procedure is by identifying outliers via the uncertainty of the model. We classified an outlier as a point in which the predicted point-wise std of our model was high. Thus, we set a threshold for the point-wise std according to the number of outliers that were added to the training set. By using this threshold, outliers were identified. For most of the datasets, both the regularized and unregularized models identified the same proportion of outliers ($\sim 10\%$). For the Allen-Cahn dataset, the regularized model outperformed the unregularized model by identifying $21.6\%$ of the outliers, whereas the unregularized model only identified $10.8\%$. The VAE-MLP method identifies $11\%$ of the outliers for the HJB dataset, and $\sim5\%$ of the outliers for the remaining four datasets. Thus, for outlier detection, PCENet performs better than VAE-MLP on four out the five datasets.

\begin{footnotesize}
   
 \begin{table}[tb!]
 \caption{\label{robust-table} Robust learning: test errors in the presence of outliers for unregularized and regularized PCE-Net, 2 layer MLP, and  VAE-MLP. }
 \begin{center}

\begin{tabular}{c|c|c|c|c|c|c|c}
\toprule
& \multicolumn{4}{c|}{Wasserstein} & \multicolumn{3}{c}{Mahalanobis}\\
\cline{2-8}
& \multicolumn{2}{c|}{PCENet} & & & \multicolumn{2}{c|}{PCENet}\\
\cline{2-8}
Datasets & Unreg & Reg & MLP & VAE-MLP & Unreg & Reg  & VAE-MLP  \\
\hline
\hline
Allen-Cahn & 0.0051 & \textbf{0.0030}  &0.1604 &0.1347 & \textbf{0.3511} & 0.3788 & 0.1646\\
\hline
HJB & 0.2192 & \textbf{0.2157} &6.1406&5.8184& \textbf{0.4428} & 0.4450 &2.3454  \\
\hline
Diabetes & 181.07 & 180.53 &144.98&\textbf{144.56}& 0.3364 & \textbf{0.3162}&  11.147\\
\hline
Parkinson &12.0621 & \textbf{11.980} &13.530&13.188& 11.5438 & \textbf{10.606}&24.929\\
\hline
Boston & 31.9227 & 27.4572  &21.043&\textbf{20.630} & 0.4431 & \textbf{0.4224} & 1.6465\\
\bottomrule
\end{tabular}

\end{center}
\end{table}

\end{footnotesize}


%
%

\section{Conclusions}

In this paper, we studied \texttt{PCE-Net}, a dimensionality reduction surrogate model-based approach for learning uncertainty in high-dimensional data systems. The method comprises two stages; namely, a dimensionality reduction stage, where VAE is used to learn a distribution of the inputs on a low-dimensional latent space, and a surrogate modeling stage, where PCE (along with an MMD regularization) are used to learn a mapping from the latent space to the output space. 
The combination of VAE and PCE provides a means to learn a functional relationship between the input and output distributions and also allows for uncertainty estimation, even when the input dimensions are high and the hidden state variables are unknown, as seen in the double-pendulum example. While the VAE can be used to generate additional training samples and ensure that the posterior distribution in the latent space captures the input distribution and dynamics, PCE along with the use of MMD as a regularizer helps to ratify that the global moments of the response match that of the outputs. In order to estimate the posterior statistics and moments it is only necessary to sample the latent space (rather than the high-dimensional input space itself), and henceforth the PCE captures the global characteristics of the data. Numerical experimental results on various datasets illustrate the utilities of the proposed method in different applications, including the robustness of the model. We observed how we can use \texttt{PCE-Net} to model system dynamics and perform UQ analysis on complex systems with high-dimensional inputs. 
We also saw that \texttt{PCE-Net} yields reasonably accurate results for supervised learning, which are comparable to standard regression techniques. It can also model uncertainty in  PDEs and learning problems.
Interesting future directions include methods for jointly learning the latent distribution and the output mapping, and directly learning the latent dimension from the data.



\bibliographystyle{siamplain}
\bibliography{pce.bib}

\newpage

\appendix

\section{Bilevel Optimization}\label{app:BAM}
In the \texttt{PCE-Net} framework with MMD regularization (Algorithm~\ref{alg:PCE_training}), the coefficients $c_k$'s of PCE and the hyperparameters $[\sigma,\lambda]$ related to the MMD regularization can be jointly optimized within a bilevel programming framework. Recently, bilevel formulation has been proven successful in hyperparameter optimization tasks~\cite{franceschi2018bilevel}, where the learning process involves two (possibly coupled) problems. The first, upper-level (UL) problem focuses on optimizing the hyperparameters using a validation dataset, while the second lower-level (LL) problem entails finding model parameters using a training dataset. However, solving such problems directly will require computing the hyper-gradient due to the chain rule and the coupling between the UL loss function and the LL optimal solution. Hence, an oracle such as Neumann approximation~\cite{lorraine2020optimizing}, capable of computing the inverse Hessian matrix of the LL loss function, will need to be used even when the LL is strongly convex. 

To alleviate this issue, more recently, penalized reformulations of the bilevel optimization problem have been explored~\cite{mehra2021penalty,shen2023penalty}. It has been analytically proven that when the penalty parameter is sufficiently large, the local and global optimal solutions for the two problems are identical, eliminating the need to compute second-order information of the loss functions. Therefore, a few different first-order based bilevel algorithms have been proposed in the literature~\cite{shen2023penalty,liu2022bome,lu2023first}, in order to overcome the computational issue of solving the bilevel optimization problems, even when the LL problem is not strongly convex. Although, the convergence behavior of these algorithms is highly dependent on the smoothness parameters of the problem. 

In our case, even though we consider the hyperparameters to be within the compact set, implying that the gradient is Lipschitz continuous, the changes in the kernel function can still be rapid locally with respect to these parameters. Targeting the issue of instability with simply plugging-in the gradient-based bilevel algorithm, we propose the proximal point based bilevel optimization framework, i.e.,  Bilevel Alternating Minimization
(BAM), for optimizing both UL and LL variables. We introduce  two surrogate functions for the UL and LL loss functions, respectively~\cite{kaplan1998proximal}. The key advantage of adopting this technique is that we can use any general optimizer as an oracle for solving each subproblem with respect to the model parameters. Particularly, when the number of hypeparameters is not large, and the nonlinearity of the loss function isthe key challenge, BFGS type algorithm \cite{wright2006numerical} will be more useful in stabilizing the convergence behavior of the learning process.




\subsection{Joint Optimization of PCE coefficients and Hyperparameters} \label{bilevel_opt}
Let $\theta$ denote the hyperparameters of \texttt{PCE-Net}, e.g., $\theta=[\sigma,\lambda], \lambda \in \mathcal{S}_\lambda,\sigma \in \mathcal{S}_\sigma$, and $\Theta$  denote the feasible set of these hyperparameters. Then, the joint hyperparameter and PCE coefficients optimization problem can be formulated as the following bilevel programming form:
\begin{subequations}
\begin{align}\label{opt.pro}
\min_{\theta\in\Theta,\{c_k\}}\quad & \mathcal{L}_{\mathrm{UL}}(\theta,\{c_k\};\mathcal{D}_\mathrm{val}) 
\\
\textrm{s.t.}&\quad  \{c_k\} \in\arg\min_{\{c'_k\}} \mathcal{L}_{\mathrm{LL}}(\theta,\{c'_k\};\mathcal{D}_{\mathrm{tr}})
\end{align}
\end{subequations}
where $\mathcal{L}_{\mathrm{UL}}$ denotes the upper-level (UL) loss function, e.g., $\mathcal{L}_{\sigma, \lambda}$, and $\mathcal{L}_{\mathrm{LL}}$ denotes the lower-level (LL) loss function, e.g., $L_2$ loss or $\mathcal{L}_{\sigma, \lambda}$.
However, as mentioned in the introduction part, solving this problem involves the computation of the hypergradient due to the nested structure of the optimization variables coupled in both UL and LL loss functions. Penalizing the LL problem to the UL objective is among the most straightforward ways of finding optimal solutions \cite{mehra2021penalty,shen2023penalty} as follows:
\begin{align*}
\min_{\theta\in\Theta,\{c_k\}}\quad & \mathcal{L}_{\mathrm{UL}}(\theta,\{c_k\};\mathcal{D}_\mathrm{val}) + \gamma p(\theta,\{c_k\};\mathcal{D}_{\mathrm{tr}}) 
\end{align*}
where $\gamma>0$, 
\begin{equation*}
    p(\theta,\{c_k\};\mathcal{D}_{\mathrm{tr}}) = \mathcal{L}_{\mathrm{LL}}(\theta,\{c_k\};\mathcal{D}_{\mathrm{tr}}) -v(\theta;\mathcal{D}_{\mathrm{tr}}),
\end{equation*}
and $v(\theta;\mathcal{D}_{\mathrm{tr}})=\min_{\{c_k\}} \mathcal{L}_{\mathrm{LL}}(\theta,\{c_k\};\mathcal{D}_{\mathrm{tr}})$ denotes the value function.

\subsubsection{Penalty-Based Bilevel Alternating Minimization (BAM)}

For sake of the notation simplicity, let $\{c^{\star}_k(\theta)\}\bydef \min_{\{c_k\}} \mathcal{L}_{\mathrm{LL}}(\theta,\{c_k\};\mathcal{D}_{\mathrm{tr}})$ and $c=[c_1,\ldots,c_{\ell_p}]$. Then, we can apply the optimization algorithm to update the upper and lower model parameters, respectively. Note that $v(\theta;\mathcal{D}_{\mathrm{tr}})$ is not dependent on $c$. Consequently, the bilevel alternating minimization algorithm can be written as follows:
\begin{subequations}
\begin{align}\notag
c_{t+1} = & \arg \min_{c\in\mathcal{C}}  \mathcal{L}_{\mathrm{UL}}(\theta_t, c ;\mathcal{D}_\mathrm{val}) 
\\
&\quad + \gamma  \mathcal{L}_{\mathrm{LL}}(\theta_t,c;\mathcal{D}_{\mathrm{tr}}) +\frac{\nu}{2}\|c -c_t\|^2,
\\\notag
\theta_{t+1}=&\arg \min_{\theta\in\Theta}  \mathcal{L}_{\mathrm{UL}}(\theta,c_{t+1};\mathcal{D}_\mathrm{val})
\\ 
 &\quad + \gamma p(\theta, c_{t+1};\mathcal{D}_{\mathrm{tr}}) 
  +\frac{\nu}{2}\|\theta -\theta_t\|^2,
\end{align}
\end{subequations}
where $t$ denotes the index of iterations.
However, $c^{\star}(\theta_{t})\in\mathcal{S}(c^{\star}(\theta_{t}))\triangleq\arg\min_{c} \mathcal{L}_{\mathrm{LL}}(\theta_t,c;\mathcal{D}_{\mathrm{tr}})$ is unknown for this problem. Towards this end, the update of $\theta$ by the BAM algorithm is as follows:
\begin{align}
\notag
\theta_{t+1}=&\arg \min_{\theta\in \Theta}  \mathcal{L}_{\mathrm{UL}}(\theta,c_{t+1};\mathcal{D}_\mathrm{val})
\\\notag
 &\quad + \gamma \left( \mathcal{L}_{\mathrm{LL}}(\theta,c_{t+1};\mathcal{D}_{\mathrm{tr}}) - \mathcal{L}_{\mathrm{LL}}(\theta, \widehat{c}_{t+1};\mathcal{D}_{\mathrm{tr}})\right) 
 \\
 & \quad +\frac{\nu}{2}\|\theta-\theta_t\|^2,
 \notag
\end{align}
where $\widehat{c}_{t+1}$ is an approximation of $c^{\star}(\theta_t)$ and can be computed by only minimizing $\mathcal{L}_{\mathrm{LL}}$.

\subsubsection{Theoretical Guarantees}

Before presenting our theoretical results, we first make the following standard assumptions and notations on the properties of this optimization problem.

{\bf Assumption 1.} Both upper-level and lower-level functions are differentiable and gradient Lipschitz continuous with constants $L_f$ and $L_g$ jointly w.r.t. $c$ and $\theta$ over the compact feasible sets $\mathcal{C}$ and $\Theta$.

\remark This assumption is standard in analyzing the dynamics of the generated sequence. As the problem is smooth and the feasible sets are compact, these properties of the loss functions hold naturally.

{\bf Assumption 2.} The value function $v(\theta;\mathcal{D}_{\mathrm{tr}})$ is differentiable with gradient Lipschitz continuous with constant $L_v$.

\remark  The loss function ${\cal L}_{\mathrm{\sigma,\lambda}}$ is smooth. Also, the function varies w.r.t. variable $\sigma$ quickly due to the sharpness of the Gaussian kernel, implying that it is unlikely that one $\theta$ corresponds to multiple solutions. So, it is reasonable to assume that the loss at $c^{\star}(\theta)$ is also smooth.

{\bf Assumption 3.} There exists an oracle such that $d_{\mathcal{S}(c^{\star}(\theta_{t}))}(\widehat{c}_{r+1})\le \delta_r$, where $d_{\mathcal{S}}(c)=\arg\min_{d'\in\mathcal{S}}\|c-d'\|$ denotes the distance between $c$ and set $\mathcal{S}$.

\remark Here, we assume that $\widehat{c}_{t+1}$
 can be obtained by applying any optimization solvers. For example, running a number of projected gradient descent steps gives
\begin{multline*}
\widehat{c}^{r+1}_t = \mathrm{proj}_{\mathcal{C}}\left( \widehat{c}^r_t - \alpha\nabla \mathcal{L}_{\mathrm{LL}}(\theta_t,\widehat{c}^r_t)\right), 
\\
\forall r=0,1\ldots, T_t-1
\end{multline*}
and $\widehat{c}_{t+1}=\widehat{c}^{T_t}_t$, where $\textrm{proj}$ denotes the projection of  $c$ within the feasible set $\mathcal{C}$. We can also apply line search type of algorithms instead so that we can find the $\delta_r$-optimal solution.

\paragraph{Convergence Rate}
Let us define the optimality gap of this problem as
\begin{equation}
\mathcal{G}(\theta_t,c_t)=\nu\left(\left[\begin{matrix} \theta_t \\ c_t\end{matrix}\right]-\mathrm{proj}\left(\left[\begin{matrix} \theta_t \\ c_t\end{matrix}\right]-\frac{1}{\nu} \left[\begin{matrix} \nabla_\theta F(\theta_t,c_t) \\ \nabla_c F(\theta_t,c_t)\end{matrix}\right]\right )\right)
\end{equation}
where
$F(\theta_t,c_t)\triangleq \gamma \left( \mathcal{L}_{\mathrm{LL}}(\theta_t,c_t ) - \mathcal{L}_{\mathrm{LL}}(\theta_t, c^{\star}(\theta_t)) \right) + \mathcal{L}_{\mathrm{UL}}(\theta_t,c_t  ). $
 It can be easily verified that $\|\mathcal{G}(\theta_t,c_t)\|=0
 $ implies that $(\theta_t,c_t)$ is a first-order stationary point of Eq. \eqref{opt.pro}.

\emph{{\bf Theorem 1.} Suppose that Assumptions 1 to 3 hold and the iterates are generated by the block-wise bilevel algorithm. When $\nu > \max\left\{3(L_f+\gamma L_g), 3L_f+\gamma (6L_g + L_v)\right\}$ and $\delta^2_r$ is summable, then the following inequality is true,
\begin{align}\notag
 &\frac{1}{T}\sum^T_{t=1}\left\|\mathcal{G}(\theta_t,c_t)\right\|^2 
 \\ \notag
& \le C\frac{  F(\theta_{1}, c_{1}) - F(\theta_{T}, c_{T})+L_g}{T}+\frac{ L^2_g\gamma^2}{T}
\end{align}
where $T$ denotes the total number of iterations, and
\begin{equation} \notag
C  \triangleq\frac{2\left( 4\nu^2+  8(L_f^2+2\gamma^2 L^2_g)\right)}{\nu}.
\end{equation}}
\begin{proof}
For simplicity of notations, let $f(\theta,c)=\mathcal{L}_{\mathrm{UL}}(\theta, c ;\mathcal{D}_\mathrm{val})$ and $g(\theta,c)=\mathcal{L}_{\mathrm{LL}}(\theta, c ;\mathcal{D}_\mathrm{tr})$. Then, the BAM algorithm can be written as
\begin{subequations}
\begin{align}
c_{t+1} = & \arg \min_{c\in\mathcal{C}}  f(\theta_t, c ) + \gamma  g(\theta_t,c ) +\frac{\nu}{2}\|c-c_t\|^2,
\\ 
\theta_{t+1}=&\arg \min_{\theta\in \Theta}  f(\theta,c_{t+1} )
 + \gamma \left( g(\theta,c_{t+1}) - g(\theta, \widehat{c}_{t+1})\right)+\frac{\nu}{2}\|\theta-\theta_t\|^2,
\end{align}
\end{subequations}
where $\widehat{c}_{t+1}$ is an approximation of $c^{\star}(\theta_t)$. 
From the optimality condition, we have
\begin{subequations}
\begin{align}
 \langle \nabla f(\theta_t, c_{t+1}) +\gamma \nabla g(\theta_t, c_{t+1}), c - c_{t+1}\rangle & \ge -\frac{\nu}{2}\|c_{t+1}-c_t\|^2, \quad \forall c\in\mathcal{C}, \label{eq.optc}
\\
 \left\langle \nabla f(\theta_{t+1}, c_{t+1}) +\gamma \left(\nabla g(\theta_{t+1}, c_{t+1}) - \nabla g(\theta_{t+1}, \widehat{c}_{t+1})\right), \theta - \theta_{t+1}\right\rangle & \ge -\frac{\nu}{2}\|\theta_{t+1} - \theta_t\|^2, \quad \forall \theta\in\Theta.
\end{align}
\end{subequations}

Note that both feasible sets $\mathcal{C}$ and $\Theta$ are compact, which gives the  Lipschitz continuity of the loss function (i.e., Assumption 1). Therefore, we obtain
\begin{align*}\notag
& F(\theta_{t},c_{t+1}) - F(\theta_{t},c_{t})
\\
\le & \langle\nabla F(\theta_t,c_t),c_{t+1}-c_t\rangle  +\frac{L_f+\gamma L_g}{2} \|c_{t+1}-c_t\|^2
\\\notag
\le & \langle\nabla f(\theta_t,c_{t+1})+\gamma \nabla g(\theta_t,c_{t+1}), c_{t+1}-c_t\rangle  +\frac{L_f+\gamma L_g}{2} \|c_{t+1}-c_t\|^2
\\
& -\langle\nabla f(\theta_t,c_{t+1})+\gamma\nabla g(\theta_t,c_{t+1}) -\nabla f(\theta_t,c_{t})-\gamma g(\theta_t,c_t)) ,c_{t+1}-c_t\rangle
 \\\notag
\le & -\left( \nu -\frac{L_f+\gamma L_g}{2}\right)\|c_{t+1}-c_{t}\|^2
\\
& -\langle\nabla f(\theta_t,c_{t+1})+\gamma\nabla g(\theta_t,c_{t+1}) -\nabla f(\theta_t,c_{t})-\gamma g(\theta_t,c_t)) ,c_{t+1}-c_t\rangle
\\
\le & -\left( \nu -\frac{3(L_f+\gamma L_g)}{2}\right)\|c_{t+1}-c_{t}\|^2
\end{align*}
where $F(\theta_t,c_t)=f(\theta_t,c_t  )
 + \gamma \left( g(\theta_t,c_t ) - g(\theta_t, c^{\star}_t)\right)$, the second inequality holds due to the optimality condition \eqref{eq.optc} by substituting $c$ by $c_t$, and the last inequality is true since we have
 \begin{subequations}
 \begin{align*}
  \langle\nabla f(\theta_t,c_{t+1})-\nabla f(\theta_t,c_{t}),c_{t+1}-c_t\rangle&\le L_f\|c_{t+1}-c_t\|^2,
  \\
    \langle\nabla g(\theta_t,c_{t+1})-\nabla g(\theta_t,c_{t}),c_{t+1}-c_t\rangle&\le L_g\|\theta_{t+1}-\theta_t\|^2.
 \end{align*}
 \end{subequations}
Similarly, we can also have
{\small
\begin{align*}\notag
& F(\theta_{t+1}, c_{t+1}) - F(\theta_{t}, c_{t+1})
\\
\le &   \langle \nabla F(\theta_r,c_{t+1}), \theta_{t+1}-\theta_t\rangle+\frac{L_f+\gamma (L_g + L_v)}{2} \|\theta_{t+1}-\theta_t\|^2
\\\notag
\le &   \langle \nabla f(\theta_{t+1}, c_{t+1}) +\gamma \left(\nabla g(\theta_{t+1}, c_{t+1}) - \nabla g(\theta_{t+1}, \widehat{c}_{t+1})\right), \theta_{t+1}-\theta_t\rangle
\\\notag
& +\frac{L_f+\gamma (L_g + L_v)}{2} \|\theta_{t+1}-\theta_t\|^2
 -\langle \nabla f(\theta_{t+1},c_{t+1}) -\nabla f(\theta_{t},c_{t+1}), \theta_{t+1}-\theta_t\rangle
\\\notag
&-\langle \gamma \left(\nabla g(\theta_{t+1}, c_{t+1}) - \nabla g(\theta_{t}, c_{t+1})\right), \theta_{t+1}-\theta_t\rangle
-\langle \gamma \left(\nabla g(\theta_{t+1}, \widehat{c}_{t+1}) - \nabla g(\sigma_{t}, \widehat{c}_{t+1})\right), \theta_{t+1}-\theta_t\rangle
\\
&-\langle \gamma \left(\nabla g(\theta_{t}, \widehat{c}_{t+1}) - \nabla g(\theta_{t}, c^{\star}(\theta_{t}))\right), \theta_{t+1}-\theta_t\rangle
\\
\le & -\left( \nu -\frac{(3L_f+\gamma (5L_g + L_v))}{2} \right)\|\theta_{t+1}-\theta_{t}\|^2-\gamma \langle  \nabla g(\theta_{t}, \widehat{c}_{t+1}) - \nabla g(\theta_{t}, c^{\star}(\theta_{t})), \theta_{t+1}-\theta_t\rangle
\\
\le & -\left( \nu -\frac{(3L_f+\gamma (6L_g + L_v))}{2} \right)\|\theta_{t+1}-\theta_{t}\|^2 +\frac{L_g}{2}d^2_{\mathcal{S}(c^{\star}(\theta_{t}))}(\widehat{c}_{t+1})
\end{align*}
}
where the first inequality holds due to the gradient Lipschitz continuity (i.e., Assumption 1), in the second inequality we apply the gradient Lipschitz continuity of $v(\theta;\mathcal{D}_{\mathrm{tr}})$ with respect to $\theta$ (i.e., Assumption 2), and the last inequality holds since
\begin{align*}
  \quad \langle  \nabla g(\theta_{t}, \widehat{c}_{t+1}) - \nabla g(\theta_{t}, c^{\star}(\theta_{t})), \theta_{t+1}-\theta_t\rangle
 \le \frac{L_g}{2}d^2_{\mathcal{S}(c^{\star}(\theta_{t}))}(\widehat{c}_{t+1}) +\frac{L_g}{2}\|\theta_{t+1}-\theta_t\|^2
\end{align*}
by applying Cauchy–Schwarz inequality.
Under the oracle assumption (i.e., Assumption 3), we obtain
{\small
\begin{align*}\notag
 &F(\theta_{t+1}, c_{t+1}) - F(\theta_{t}, c_{t})
 \\
 &\le -\left( \nu -\frac{3(L_f+\gamma L_g)}{2}\right)\|c_{t+1}-c_{t}\|^2-\left( \nu -\frac{(3L_f+\gamma (6L_g + L_v))}{2} \right)\|\theta_{t+1}-\theta_{t}\|^2 +\frac{L_g\delta^2_t}{2}
  \\
 &\le - \frac{\nu}{2}\|c_{t+1}-c_{t}\|^2-  \frac{\nu}{2} \|\theta_{t+1}-\theta_{t}\|^2 +\frac{L_g\delta^2_t}{2}
 \end{align*}
where the second inequality holds for
\begin{align*}
\nu > \max\left\{3(L_f+\gamma L_g), 3L_f+\gamma (6L_g + L_v)\right\}.
\end{align*}
}
Applying the telescoping sum gives
\begin{align} \label{eq.tel}
  \frac{\nu}{2}\frac{1}{T}\sum^T_{t=1}\left(\|c_{t+1}-c_{t}\|^2+ \|\theta_{t+1}-\theta_{t}\|^2 \right)
 \le \frac{F(\theta_{1}, c_{1}) - F(\theta_{T}, c_{T})+L_g}{T} ,
\end{align}
where we use the fact that $\sum^T_{t=1}t^{-2}\le1+\int^T_{1}x^{-2}dx=2-T^{-1}$ for $\delta_r=1/t$ (i.e., when $\delta^2_r=1/t^2$ is summable).
 
Let us define the optimality gap as
\begin{equation*}
\mathcal{G}(\theta_t,c_t)=\nu\left(\left[\begin{matrix} \theta_t \\ c_t\end{matrix}\right]-\textrm{proj}\left(\left[\begin{matrix} \theta_t \\ c_t\end{matrix}\right]-\frac{1}{\nu} \left[\begin{matrix} \nabla_{\theta} F(\theta_t,c_t) \\ \nabla_c F(\theta_t,c_t)\end{matrix}\right]\right )\right).
\end{equation*}

Thus,
\begin{align}\notag \label{eq.proj}
&\quad\left\|\mathcal{G}(\theta_t,c_t)\right\|
\\
&\le \nu\|z_t-z_{t+1}\| + \nu\left\|z_{t+1} -\textrm{proj}\left(\left[\begin{matrix} \theta_t \\ c_t\end{matrix}\right]-\frac{1}{\nu} \left[\begin{matrix} \nabla_{\theta} F(\theta_t,c_t) \\ \notag \nabla_c F(\theta_t,c_t)\end{matrix}\right]\right ) \right\|
\\ 
&\le 2\nu\|z_t-z_{t+1}\| + \left\|\left[\begin{matrix} \nabla_{\theta} \widehat{F}(\theta_{t+1},c_{t+1}) \\ \nabla_c \widehat{F}(\theta_{t+1},c_{t+1})\end{matrix}\right] - \left[\begin{matrix} \nabla_{\theta} F(\theta_t,c_t) \\ \nabla_c F(\theta_t,c_t)\end{matrix}\right] \right\|
\end{align}
Define
\begin{equation*}
z_t=\left[\begin{matrix} \theta_t \\ c_t\end{matrix}\right] .
\end{equation*}
Then, the second inequality of \eqref{eq.proj} holds since
\begin{equation*}
z_{t+1} = \textrm{proj}\left(\left[\begin{matrix} \theta_{t+1} \\ c_{t+1}\end{matrix}\right]-\frac{1}{\nu} \left[\begin{matrix} \nabla_{\theta} \widehat{F}(\theta_{t+1},c_{t+1}) \\ \nabla_c \widehat{F}(\theta_{t+1},c_{t+1})\end{matrix}\right]\right ),
\end{equation*}
and also $\widehat{F}(\theta_t,c_t)=f(\theta_t,c_t  )
 + \gamma \left( g(\theta_t,c_t ) - g(\theta_t, \widehat{c}_t)\right)$. Therefore, we obtain
 \begin{align*}\notag
&\quad \left\|\mathcal{G}(\theta_t,c_t)\right\|^2
\\
&\le \left(4\nu^2+8(L^2_f+2\gamma^2 L^2_g)\right)\|z_{t+1}-z_t\|^2+ 8\gamma^2d^2_{\mathcal{S}(c^{\star}(\theta_t))}( \widehat{c}_{t+1}) ,
 \end{align*} 
where we use
{\small
 \begin{align*}\notag
 &\quad \left\|\left[\begin{matrix} \nabla_\theta \widehat{F}(\theta_{t+1},c_{t+1}) \\ \nabla_c \widehat{F}(\theta_{t+1},c_{t+1})\end{matrix}\right] - \left[\begin{matrix} \nabla_\theta F(\theta_t,c_t) \\ \nabla_c F(\theta_t,c_t)\end{matrix}\right] \right\|^2
 \\
 &\le \|\nabla_\theta \widehat{F}(\theta_{t+1},c_{t+1})  - \nabla_\theta F(\theta_t,c_t) \|^2 + \|\nabla_c \widehat{F}(\theta_{t+1},c_{t+1}) -\nabla_c F(\theta_t,c_t)\|^2
 \\\notag
 &\le\left\| \nabla f(\theta_{t+1}, c_{t+1}) +\gamma \nabla g(\theta_{t+1}, c_{t+1}) -\left(\nabla f(\theta_t, c_{t}) +\gamma \nabla g(\theta_t, c_{t})\right)\right\|^2
 \\\notag
 &\quad +\bigg\|\nabla f(\theta_{t+1}, c_{t+1}) +\gamma \left(\nabla g(\theta_{t+1}, c_{t+1}) - \nabla g(\theta_{t+1}, \widehat{c}_{t+1})\right)
 \\
  &\quad - \left(\nabla f(\theta_{t}, c_{t}) +\gamma \left(\nabla g(\theta_{t}, c_{t}) - \nabla g(\theta_{t}, {c}^{\star}(\theta_{t}))\right)\right)\bigg\|^2
  \\
  &\le 4(L^2_f+\gamma^2 L^2_g)\|z_{t+1}-z_t\|^2+2\gamma^2\|\nabla g(\theta_{t+1}, \widehat{c}_{t+1}) - \nabla g(\theta_{t}, \widehat{c}_{t+1})+ \nabla g(\theta_{t}, \widehat{c}_{t+1}) - \nabla g(\theta_{t}, {c}^{\star}(\theta_{t}))\|^2
\\
  &\le 4(L^2_f+\gamma^2 L^2_g)\|z_{t+1}-z_t\|^2+4\gamma^2L^2_g\| \theta_{t+1} - \theta_{t}\|^2+ 4\gamma^2\|\nabla g(\theta_{t}, \widehat{c}_{t+1}) - \nabla g(\theta_{t}, {c}^{\star}(\theta_{t}))\|^2 
  \\
  &\le 4(L^2_f+2\gamma^2 L^2_g)\|z_{t+1}-z_t\|^2+ 4\gamma^2d^2_{\mathcal{S}(c^{\star}(\theta_t))}( \widehat{c}_{t+1}).
  \end{align*}
}
 Combining \eqref{eq.tel} yields
 \begin{equation*}
 \frac{1}{T}\sum^T_{t=1}\left\|\mathcal{G}(\theta_t,c_t)\right\|^2\le\frac{2\left( 4\nu^2+  8(L_f^2+2\gamma^2 L^2_g)\right)}{\nu}\frac{  F(\theta_{1}, c_{1}) - F(\theta_{T}, c_{T})+L_g}{T}+\frac{ L^2_g\gamma^2}{T}.
 \end{equation*}

\end{proof}

\remark The result implies that $\min_{t}\|\mathcal{G}(\theta_t,c_t)\|^2\le\mathcal{O}(1/T)$ and the convergence rate of achieving the $\epsilon$-stationary points of this problem is $\mathcal{O}(1/\epsilon^2)$, where $(\theta^{\star},c^{\star})$ that is an $\epsilon$-stationary point satisfies
$ \|\mathcal{G}(\theta^{\star},c^{\star})\|\le\epsilon.
$ 
\begin{figure*}[!tb]
\centering{}%
\includegraphics[clip,width = 5 cm]{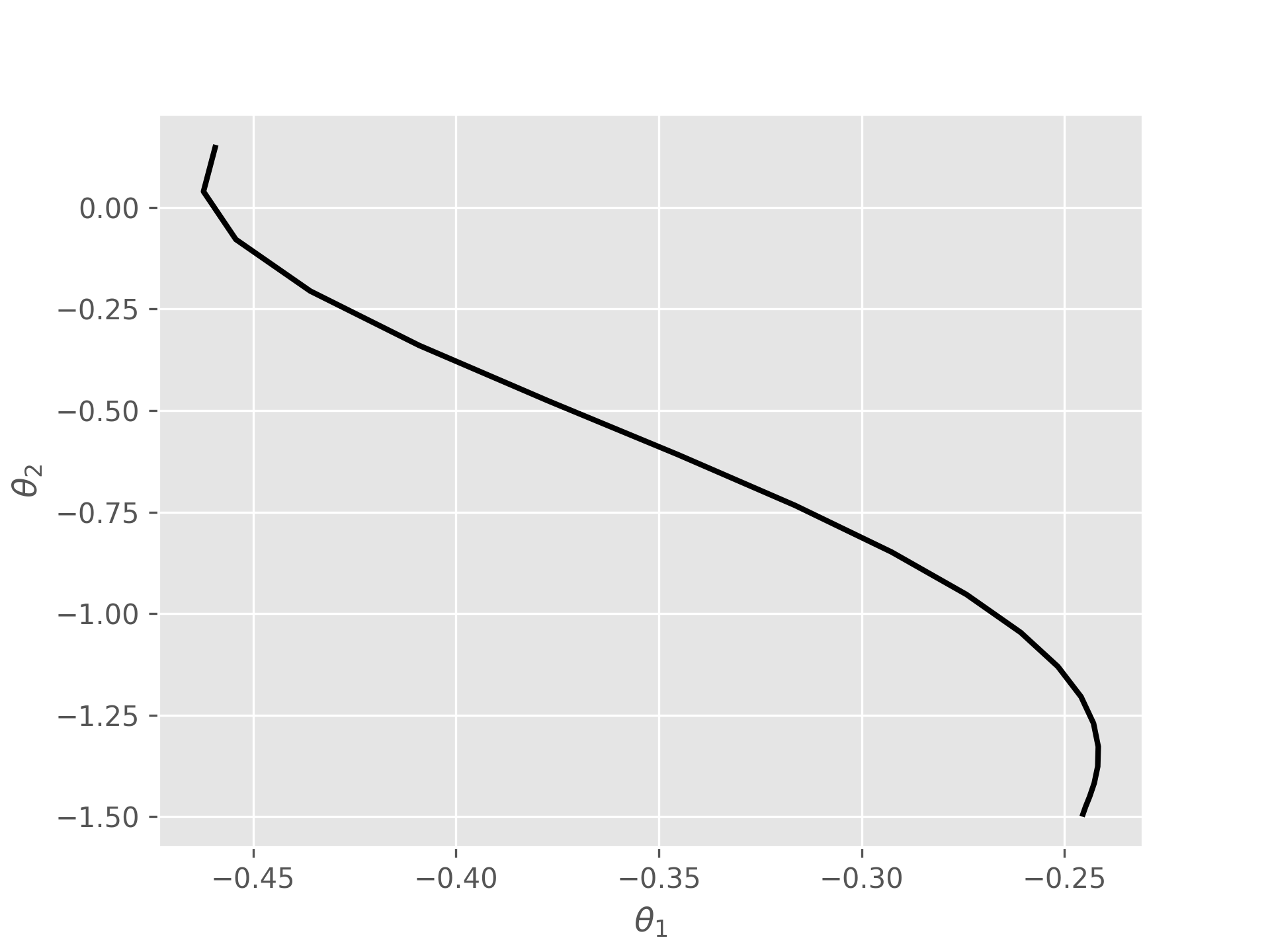} 
\includegraphics[clip,width = 5 cm]{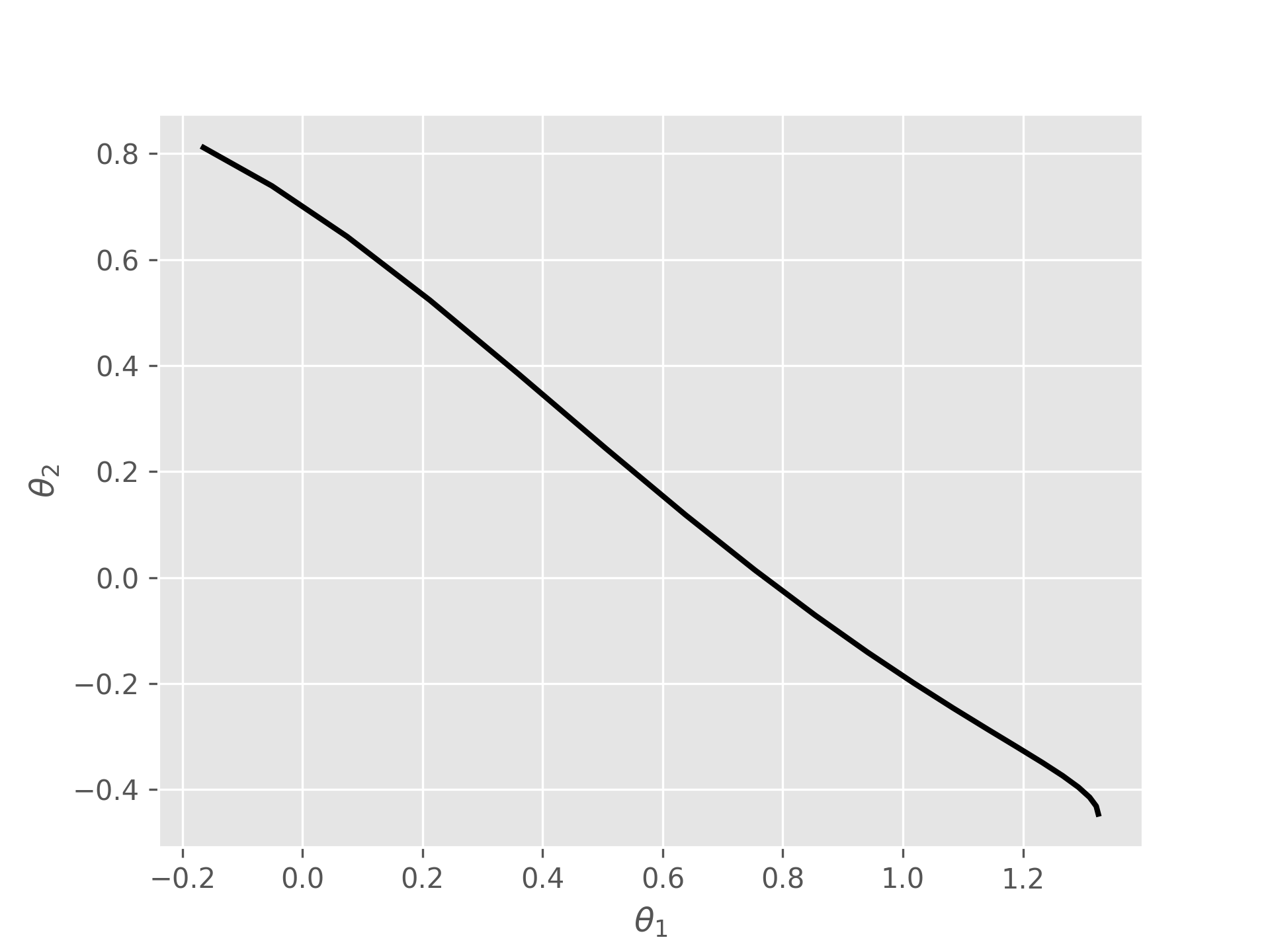} 
\includegraphics[clip,width = 5 cm]{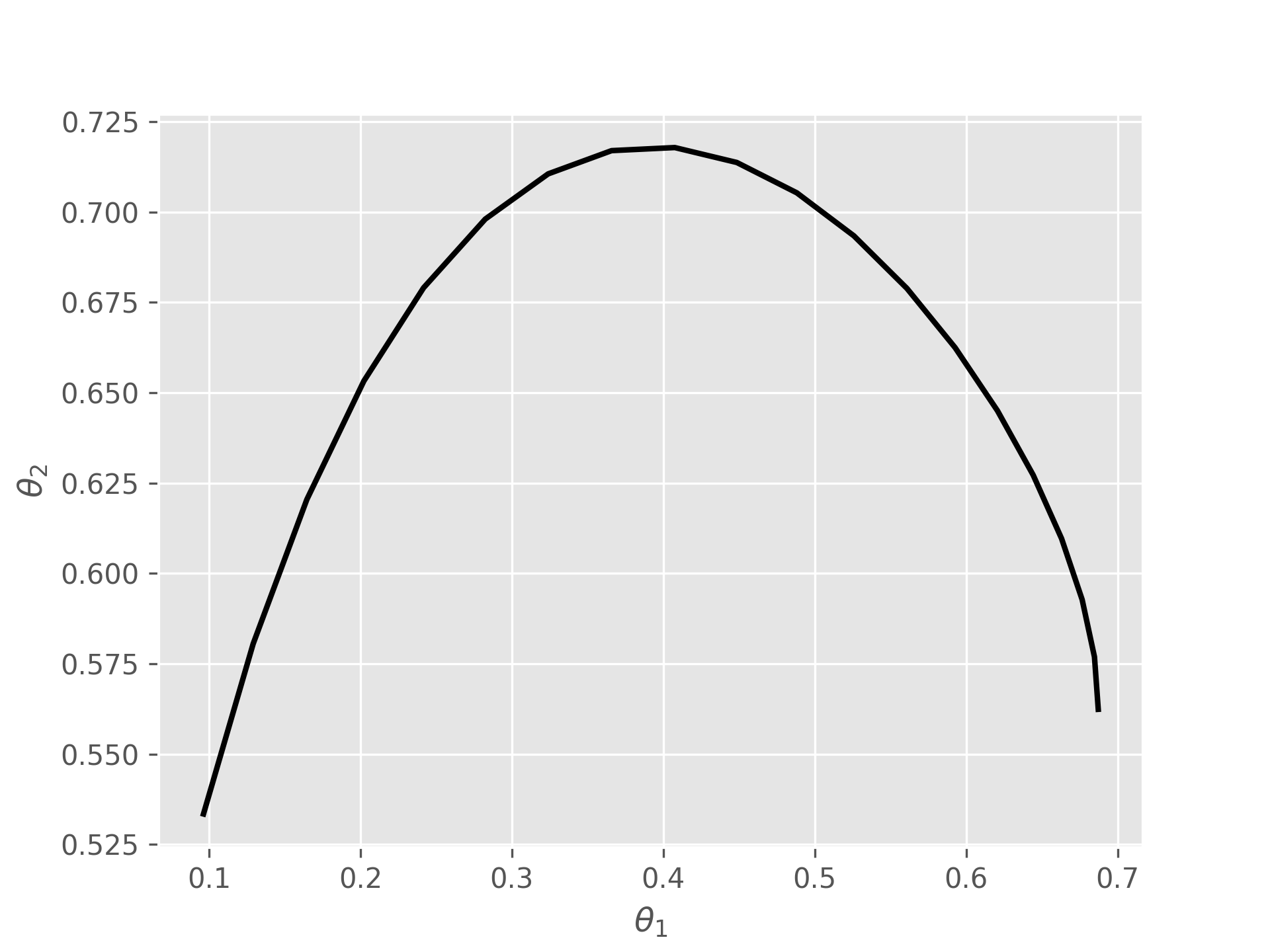}
\includegraphics[clip,width = 5 cm]{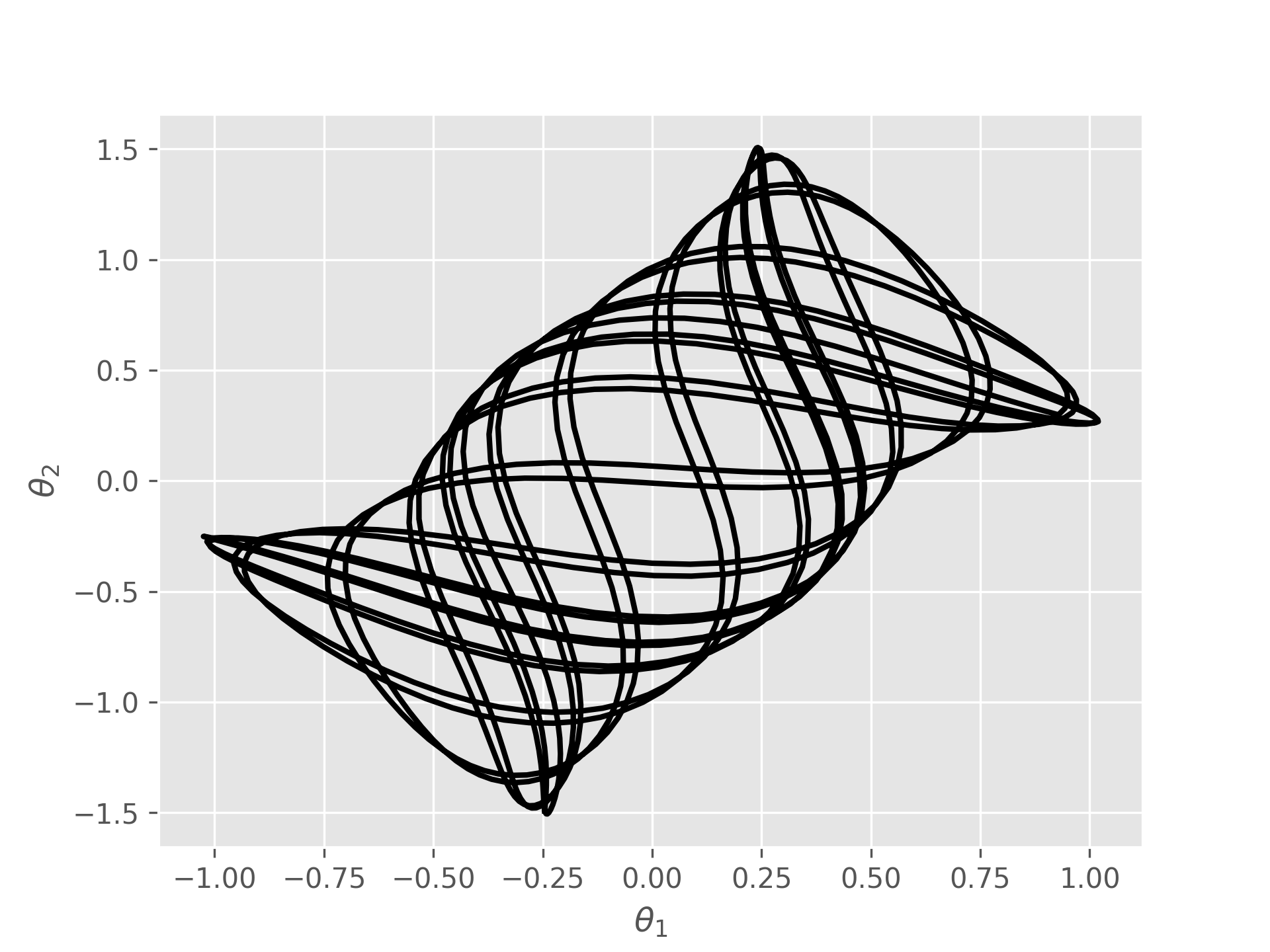}
\includegraphics[clip,width = 5 cm]{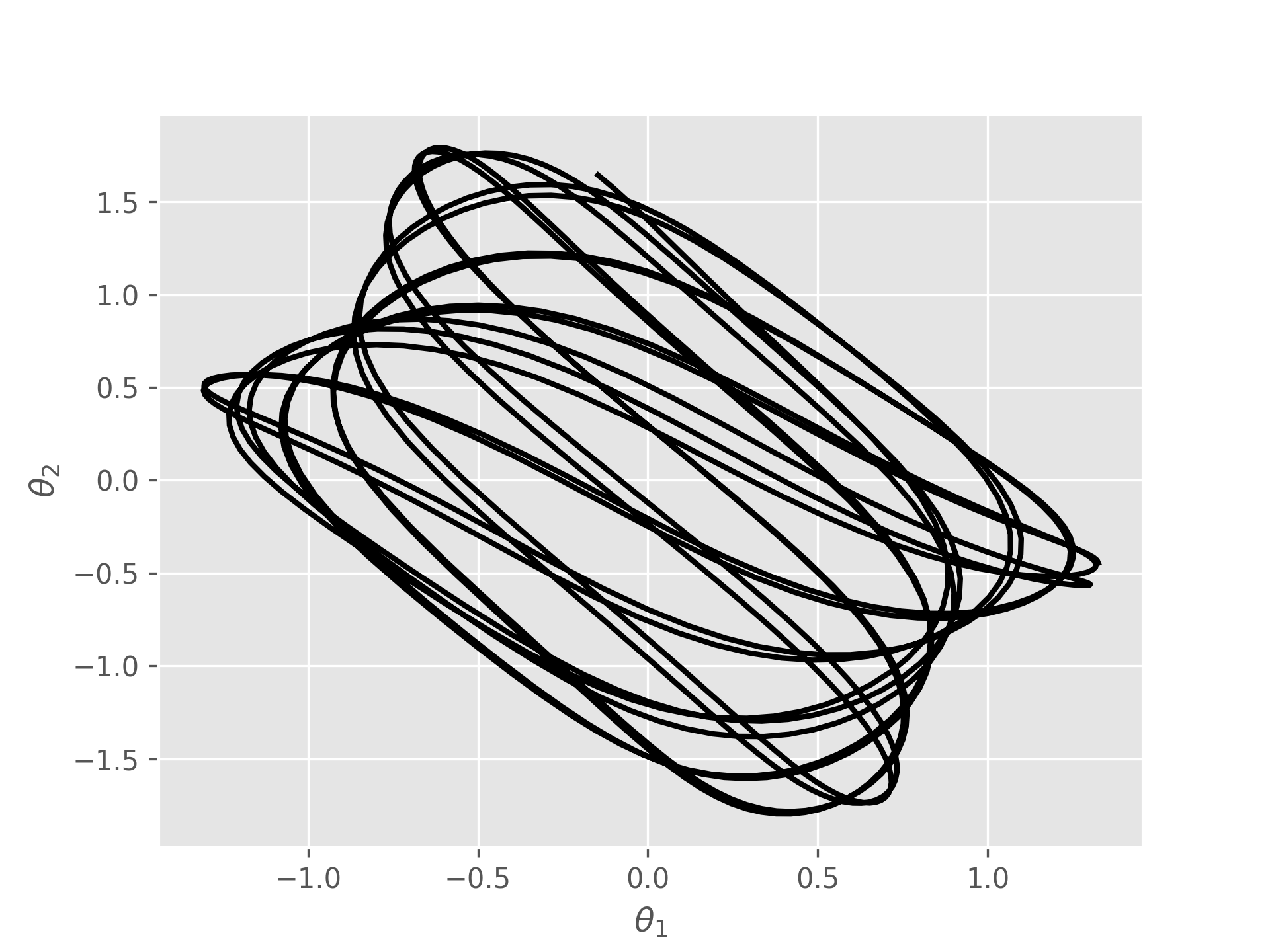}
 \includegraphics[clip,width = 5 cm]{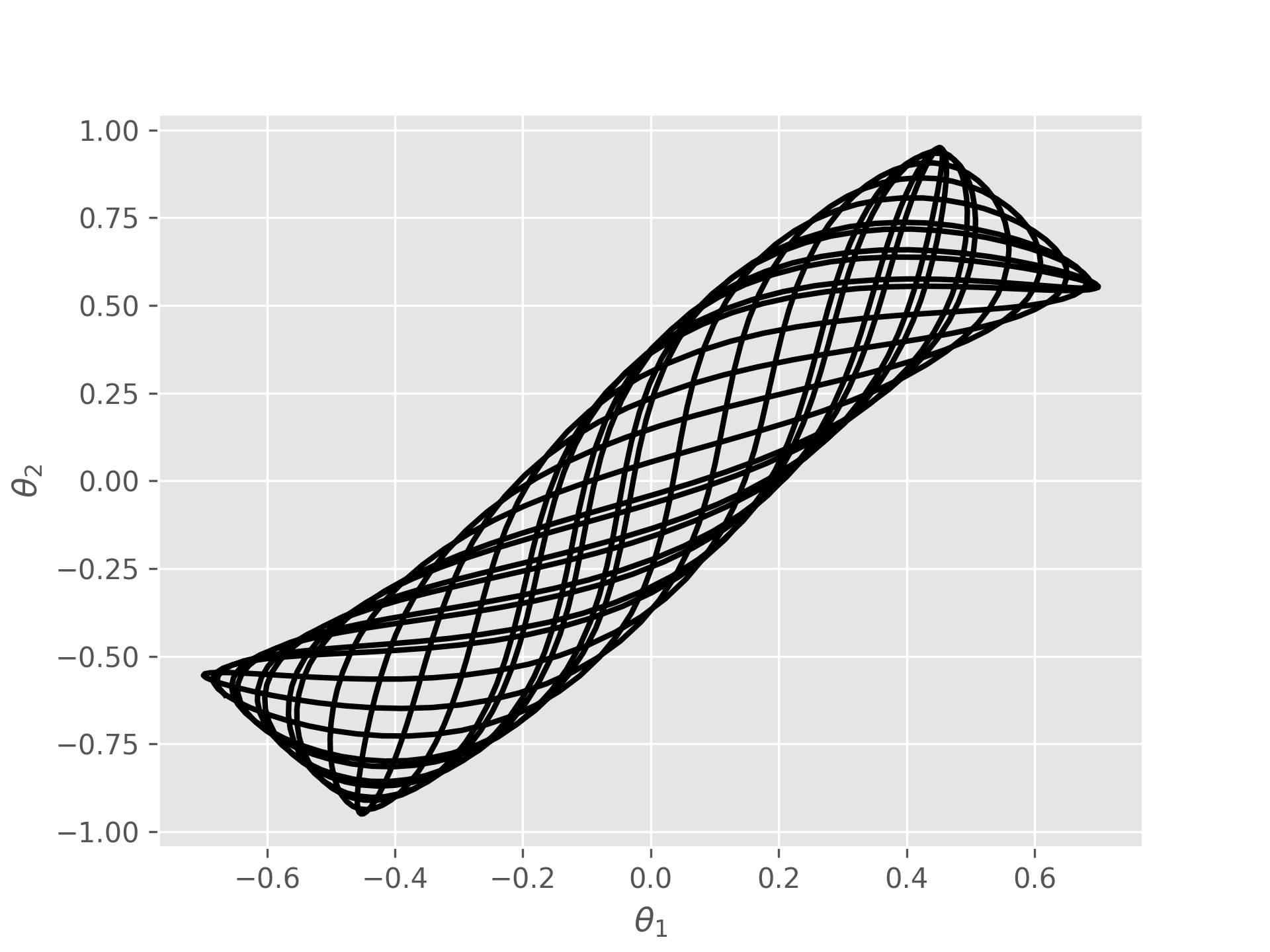}
\caption{\label{DP-fig} Trajectories (Dynamics) of the two parameters ($\theta_1$ and $\theta_2$) in the double pendulum problem. Top: Three example  trajectories for 20 time steps, and  Bottom: same examples ran for 1000 time steps.}
\vskip -0.1in
\end{figure*}
\section{Additional results and details}
\subsection{Dynamics of double pendulum}\label{app:dp}
The numerical solution of the double pendulum equations produces a complex interplay between the two parameters, i.e., the angles of the first mass $\theta_1$ and the second mass $\theta_2$, respectively. The problem is sensitive to the initial conditions. Different trajectories are generated for different realizations of the starting angles of each rods. Three such examples of the trajectory projections in configuration space are shown in Fig.~\ref{DP-fig} for the same initial momentum but different initial angles. In the plots, the X and the Y-axes are  the two angles $\theta_1$ and $\theta_2$, respectively. The top three plots give us the trajectory for 20 time steps, while the bottom three are for the same systems ran for 1000 time steps. In the paper, we use as the output the position of the rods at $t=20\Delta t$. In order to visualize the dynamics, we show the respective trajectories for longer times in bottom plots of the figure.

\subsection{Additional details for robust learning}\label{app:RL}

The parameters found using grid search can be found in Table \ref{robust-hyper-table}.

 

\begin{table}[H]
\caption{\label{robust-hyper-table} Robust learning: hyperparameters.}
\begin{center}
\begin{tabular}{ | c | c|c| }
\hline
  Dataset & $\lambda$ &  $\sigma$ \\ 
  \hline
  Allen-Cahn &  $0.1$ & $0.1$ \\
  \hline
  HJB & $0.1$ & $\sqrt{0.05}$ \\
  \hline
  Diabetes & $1$ & $\sqrt{0.1}$ \\
  \hline
  Parkinson & $10$ & $0.1$ \\
  \hline
  Boston & $0.01$ & $\sqrt{0.5}$ \\
  \hline
\end{tabular}
\end{center}
\end{table}

\end{document}